\documentclass{article}

\PassOptionsToPackage{numbers,square,comma}{natbib}

\usepackage[final]{neurips_2020}

\usepackage[utf8]{inputenc} %
\usepackage[T1]{fontenc}    %
\usepackage{hyperref}       %
\usepackage{url}            %
\usepackage{booktabs}       %
\usepackage{amsfonts}       %
\usepackage{nicefrac}       %
\usepackage{microtype}      %

\usepackage{amsmath}
\usepackage{amssymb}
\usepackage{interval}
\usepackage{xcolor}
\usepackage{subcaption}
\usepackage{algorithm}
\usepackage{algorithmic}
\usepackage{multirow}
\usepackage{enumitem}
\usepackage{graphicx}
\usepackage{wrapfig}

\graphicspath{{./}{figs/}}

\newcommand{\problem}{NCFNL}
\newcommand{\method}{MetaTNE}

\newcommand{\CG}{\mathcal{G}}
\newcommand{\CV}{\mathcal{V}}
\newcommand{\CE}{\mathcal{E}}
\newcommand{\CY}{\mathcal{Y}}
\newcommand{\CD}{\mathcal{D}}
\newcommand{\CT}{\mathcal{T}}
\newcommand{\CS}{\mathcal{S}}
\newcommand{\CQ}{\mathcal{Q}}
\newcommand{\CL}{\mathcal{L}}
\newcommand{\CN}{\mathcal{N}}

\newcommand{\SL}{\ell}

\newcommand{\V}[1]{\mathbf{#1}}
\newcommand{\M}[1]{\mathbf{#1}}

\newcommand{\dist}[2]{\mathrm{dist}(#1,#2)}

\newcommand{\dotproduct}{\cdot}
\newcommand{\authorspace}{,\ }

\title{Node Classification on Graphs with Few-Shot Novel Labels via Meta Transformed Network Embedding}
\author{%
Lin Lan$^{1}$\thanks{\,Lin, Pinghui, and Xuefeng contributed equally to this work.}\authorspace\ 
Pinghui Wang$^{1}$\footnotemark[1]\protect\phantom{$^{\ast}$}\thanks{\,Pinghui Wang is the corresponding author.}\authorspace\ 
Xuefeng Du$^{2}$\footnotemark[1]\authorspace\ 
Kaikai Song$^{3}$\authorspace
Jing Tao$^{1}$\authorspace
Xiaohong Guan$^{1,4}$\\
$^1$\,MOE Key Laboratory of Intelligent Networks and Network Security,\\School of Automation Science and Engineering, Xi'an Jiaotong University, China\\
$^2$\,University of Wisconsin-Madison, USA \hspace{2mm}
$^3$\,Huawei Noah's Ark Lab, China\\
$^4$\,Department of Automation and NLIST Lab, Tsinghua University, China\\
\texttt{llan@sei.xjtu.edu.cn},\;\;
\texttt{\{phwang, jtao, xhguan\}@mail.xjtu.edu.cn}\\
\texttt{xdu66@wisc.edu},\;\;
\texttt{caesar.song@huawei.com}
}

\begin{document}

\maketitle

\begin{abstract}
We study the problem of node classification on graphs with few-shot novel labels,
which
has two distinctive properties:
(1) There are novel labels to emerge in the graph;
(2) The novel labels have only a few representative nodes for training a classifier.
The study of this problem is instructive and corresponds to many applications
such as recommendations for newly formed groups with only a few users in online social networks.
To cope with this problem,
we propose a novel Meta Transformed Network Embedding framework (MetaTNE),
which consists of three modules:
(1) A \emph{structural module} provides each node a latent representation according to the graph structure.
(2) A \emph{meta-learning module} captures the relationships between the graph structure and the node labels as prior knowledge in a meta-learning manner.
Additionally, we introduce an \emph{embedding transformation function} that remedies the deficiency of the straightforward use of meta-learning.
Inherently, the meta-learned prior knowledge can be used to facilitate the learning of few-shot novel labels.
(3) An \emph{optimization module} employs a simple yet effective scheduling strategy to train
the above two modules with a balance between graph structure learning and meta-learning.
Experiments on four real-world datasets
show that \method{} brings a huge improvement over the state-of-the-art methods.

\end{abstract}

\section{Introduction}\label{sec:intro}
Graphs are ubiquitously used to represent data in a wide range of fields,
including social network analysis, bioinformatics, recommender systems, and computer network security.
Accordingly, graph analysis tasks, such as node classification, link prediction, and community detection,
have a significant impact on our lives in reality.
In this paper, we focus on the task of node classification.
Particularly, we consider the classification of \emph{few-shot novel labels},
which means
there are some novel labels to emerge in the graph of interest
and the novel labels usually have only a few representative nodes
including the positive and the negative (i.e., holding and not holding the novel labels, respectively).
Hereafter, we refer to the \emph{available positive and negative nodes} of a label
as the \emph{support nodes} of that label.
The study of \emph{Node Classification on graphs with Few-shot Novel Labels} (\problem{})
is instructive for many practical applications.
Let us consider the following scenarios.

\noindent
\textbf{Motivating Examples}.
(1)
Some organizations in online social networks, such as Facebook, Twitter, and Flickr,
may distribute advertisements about whether users are interested in their new features
or are willing to join their new social media groups.
Through \problem{}, these organizations can predict other users' preferences
based on positive and negative responses of a few users
and provide better services or recommendations without too much bother for users.
(2) For biological protein-protein networks,
some researchers may discover a new biological function of certain proteins.
Given a few proteins with and without a specific function,
the study of \problem{} could predict whether other proteins have the function,
which helps recommend new directions for wet laboratory experimentation.

Some straightforward ways
could be derived
from existing unsupervised or semi-supervised network embedding methods
while suffer from low performance,
and please refer to \S~\ref{sec:related} for detailed discussions.
To tackle this problem,
we argue that
different labels in a graph share some intrinsic evolution patterns (e.g., the way a label propagates along the graph structure according to the proximities between nodes).
Assuming that there are a set of labels that have sufficient support nodes
(e.g., interest groups that have existed and evolved for a long time in online social networks
and protein functions that biologists are already familiar with),
we desire to extract the common patterns from the graph structure and these labels
and then utilize the found patterns to help recognize few-shot novel labels.
However, the relationships between the graph structure and node labels are complex and there could be various propagation patterns between nodes.
It remains challenging to design a model to capture all the patterns,
and how to apply them to novel labels still needs to be further studied.

\noindent
\textbf{Overview of Our Approach.}
Inspired by recent advances in few-shot learning through meta-learning~\cite{ravi2016optimization,finn2017model,snell2017prototypical,lan2019meta},
we cast the problem of \problem{} as a meta-learning problem
and propose a novel Meta Transformed Network Embedding framework, namely MetaTNE,
which allows us to exploit the common patterns.
As shown in Fig.~\ref{fig:framework},
our proposed framework consists of three modules:
\emph{the structural module}, \emph{the meta-learning module}, and \emph{the optimization module}.
Given a graph and a set of labels (called known labels) with sufficient support nodes,
the structural module first learns a latent representation for each node according to the graph structure.
Then,
considering that we ultimately expect to recognize few-shot novel labels, we propose the meta-learning module to simulate the few-shot scenario during the training phase
instead of directly performing optimization over all known labels.
Moreover, most existing meta-learning works~\cite{finn2017model,snell2017prototypical}
focus on image- and text-related tasks, while the graph structure is more irregular in nature.
To adequately exploit the complex and multifaceted relationships between nodes,
we further design an \emph{embedding transformation function} to map the structure-only (or task-agnostic) node representations to the task-specific ones
for different few-shot classification tasks.
To some extent, the meta-learning module implicitly encodes the shared propagation patterns of different labels
through learning a variety of tasks.
Finally,
the optimization module is proposed to train the preceding two modules with a simple yet effective scheduling strategy
in order to ensure the training stability and the effectiveness.
One advantage of \method{} is that, after training,
it is natural to directly apply the learned meta-learning module to
few-shot novel labels.

Our main contributions are summarized as follows:
\begin{itemize}[leftmargin=*]
\item
We explore to only use the graph structure and some known labels to
study the problem of \problem{}.
Compared with previous graph convolution based works~\cite{zhou2019meta,yao2019graph} that rely on high-quality node content for feature propagation and aggregation,
our work is more challenging and at the same time more applicable to content-less scenarios.
\item
We propose an effective framework
to solve \problem{} in a meta-learning manner. Our framework
is able to generalize to classifying emerging novel labels with only a few support nodes.
In particular,
we design a transformation function that captures the multifaceted relationships between nodes to facilitate applying meta-learning to the graph data.
\item
We conduct extensive experiments on four publicly available real-world datasets,
and empirical results show that \method{}
achieves up to 150.93\% and 47.58\% performance improvement
over the state-of-the-art methods in terms of Recall
and F$_1$, respectively.
\end{itemize}

\section{Related Work}\label{sec:related}
\noindent
\textbf{Unsupervised Network Embedding.}
This line of works focus on learning node embeddings that preserve various structural relations between nodes~\cite{zhang2018network,bonner2019exploring},
including skip-gram based methods~\cite{perozzi2014deepwalk,tang2015line,grover2016node2vec,DBLP:conf/kdd/RibeiroSF17},
deep learning based methods~\cite{cao2016deep,wang2016structural},
and matrix factorization based methods~\cite{cao2015grarep,qiu2018network}.
A straightforward way to adapt these methods for \problem{}
is to simply train a new classifier (e.g., logistic regression)
when novel labels emerge, while the learned node embeddings hold constant.
However, this does not incorporate the guidance from node labels
into the process of network embedding,
which dramatically degrades the performance in the few-shot setting.

\noindent
\textbf{Semi-Supervised Network Embedding.}
These approaches typically formulate a unified objective function to jointly optimize the learning of node embeddings and the classification of nodes,
such as combining the objective functions of DeepWalk and support vector machines~\cite{li2016discriminative,tu2016max},
as well as
regarding labels as a kind of context and using node embeddings to simultaneously
predict structural neighbors and node labels~\cite{chen2016incorporate,yang2016revisiting}.
Another line of works~\cite{kipf2016semi,hamilton2017inductive,velivckovic2017graph,hu2019strategies,xu2019mr}
explore graph neural networks to solve semi-supervised node classification as well as graph classification.
Two recent works~\cite{li2018deeper,zhang2019fewshot} extend graph convolutional network (GCN)~\cite{kipf2016semi}
to accommodate to the few-shot setting.
However, the above methods are limited to a fixed set of labels
and the adaptation of them to \problem{}
requires to train the corresponding classification models or parameters from scratch when a novel label appears,
which is not a well-designed solution to the few-shot novel labels and usually cannot reach satisfactory performance.
Recently, \citet{DBLP:conf/iclr/ChauhanNK20}
study few-shot graph classification with unseen novel labels based on graph neural networks. \citet{DBLP:journals/corr/abs-1911-11298} propose a few-shot knowledge graph completion method that essentially performs link prediction in a novel graph given a few training links.
In comparison,
we study node classification with respect to few-shot novel labels in the same graph
and their methods are not applicable.

In addition, GCN based methods \textbf{heavily rely on} high-quality node content for
feature propagation and aggregation,
while in some networks (e.g., online social networks),
some nodes (e.g., users) may not expose or expose noisy (low-quality) content,
or even all node content is unavailable due to privacy concerns~\cite{zhang2018sine,lan2020improving},
which would limit the practical use of these methods.
In contrast,
our focus is to solve the problem of \problem{} by exploiting the relationships between the graph structure and the node labels,
without involving node content.

\noindent
\textbf{Meta-Learning on Graphs.}
\citet{zhou2019meta} propose Meta-GNN that
applies MAML~\cite{finn2017model} to GCN in a meta-learning way.
More recently, \citet{yao2019graph} propose a method that combines GCN with metric-based meta-learning~\cite{snell2017prototypical}.
To some extent, all methods could handle novel labels emerging in a graph.
However,
they are built upon GCN and thus need high-quality node content for better performance,
while in this paper we are interested in graphs without node content.

\noindent
\textbf{Few-Shot Learning on Images.}
Recently, few-shot learning has received considerable attention.
Most works~\cite{ravi2016optimization,finn2017model,snell2017prototypical,DBLP:conf/iclr/YaoWTLDLL20,DBLP:conf/iclr/LiuLPKYHY19}
focus on the problem of few-shot image classification
in which there are no explicit relations between images. Some works also introduce task-specific designs for better generalization and learnability, such as task-specific
null-space projection \cite{DBLP:conf/icml/YoonSM19} and infinite mixture prototypes \cite{DBLP:conf/icml/AllenSST19}.
However, graph-structured data exhibits complex relations between nodes (i.e., the graph structure)
which are the most fundamental and important information in a graph,
making it difficult to directly apply these few-shot methods to graphs.
In addition,
\citet{liu2019learning} propose to construct a graph of image classes and learn to
propagate messages between prototypes of different classes according to the graph structure,
of which the goal is to obtain better class prototypes for few-shot image classification.
Although this work introduces the concept of graph meta-learning, it is not applicable to our scenario where a label can be positive or negative for different nodes.

\section{Problem Formulation}\label{sec:problem}
Throughout the paper,
we use lowercase letters to denote scalars (e.g., $\SL$),
boldface lowercase letters to denote vectors (e.g., $\V{u}$),
and boldface uppercase letters to denote matrices (e.g., $\M{W}$).

We denote a graph of interest by $\CG=(\CV,\CE,\CY)$,
where $\CV=\{v_1,v_2,\ldots,v_{|\CV|}\}$ is the set of nodes,
$\CE=\{e_{ij}=(v_i,v_j)\}\subseteq\CV\times\CV$ is the set of edges,
and $\CY$ is the set of labels associated with nodes in the graph.
Here, we consider the multi-label setting where each node may have multiple labels.
Let $\SL_{v_i,y}\in\{0,1\}$ be the label indicator of the node $v_i$
in terms of the label $y\in\CY$, where $\SL_{v_i,y}=1$ suggests
that the node $v_i$ holds the label $y$ and $\SL_{v_i,y}=0$ otherwise.
We use
$\CD_{y}^{+}=\{v_i|\SL_{v_i,y}=1\}$ to denote
nodes that hold the label $y$,
and $\CD_{y}^{-}=\{v_i|\SL_{v_i,y}=0\}$ to denote
nodes that do not hold the label $y$.
In this paper, we assume $\CG$ is undirected for ease of presentation.

\noindent
\textbf{Known Labels and Novel Labels.}
We divide the labels into two categories:
the known labels $\CY_{\text{known}}$ and the novel labels $\CY_{\text{novel}}$.
The former are given before we start any kind of learning process
(e.g., semi-supervised network embedding),
while the latter emerge after we have learned a model.

We assume that
each known label is complete,
namely $|\CD_{y}^{+}|+|\CD_{y}^{-}|=|\CV|$ for $y\in\CY_{\text{known}}$.
To some extent, the known labels refer to relatively stable labels (e.g., an interest group that has existed and evolved for a long time in online social networks).
Although for some nodes, inevitably we are not sure whether they hold specific known labels or not, we simply assume that the corresponding label indicators equal $0$ (i.e., not holding) like many other node classification works~\cite{perozzi2014deepwalk,tang2015line}.
In practice, a more principled way is to additionally consider the case of uncertain node-label pairs and define the label indicator as 1, 0, and -1 for the cases of holding the label, uncertain label, and not holding the label, respectively,
which we leave as future work.

On the other hand, a novel label has only a few support nodes (e.g., $10$ positive nodes and $10$ negative nodes).
By leveraging the known labels
that have sufficiently many positive and negative nodes,
we aim to explore the propagation patterns of labels along the graph structure
and learn a model that generalizes well to classifying emerging novel labels with only a few support nodes.

\noindent
\textbf{Our Problem.}
Given a graph $\CG=(\CV,\CE,\CY_{\text{known}},\CY_{\text{novel}})$,
the problem of \problem{} aims to explore the relationships between the graph structure and the known labels $\CY_{\text{known}}$
and learn a generalizable model for classifying novel labels $\CY_{\text{novel}}$.
Specifically, for each $y\in\CY_{\text{novel}}$,
after observing only a few corresponding support nodes,
the model should be able to generate or act as a good classifier to determine whether other nodes hold the label $y$ or not.

\begin{figure*}[t]
\centering
\includegraphics[width=\linewidth]{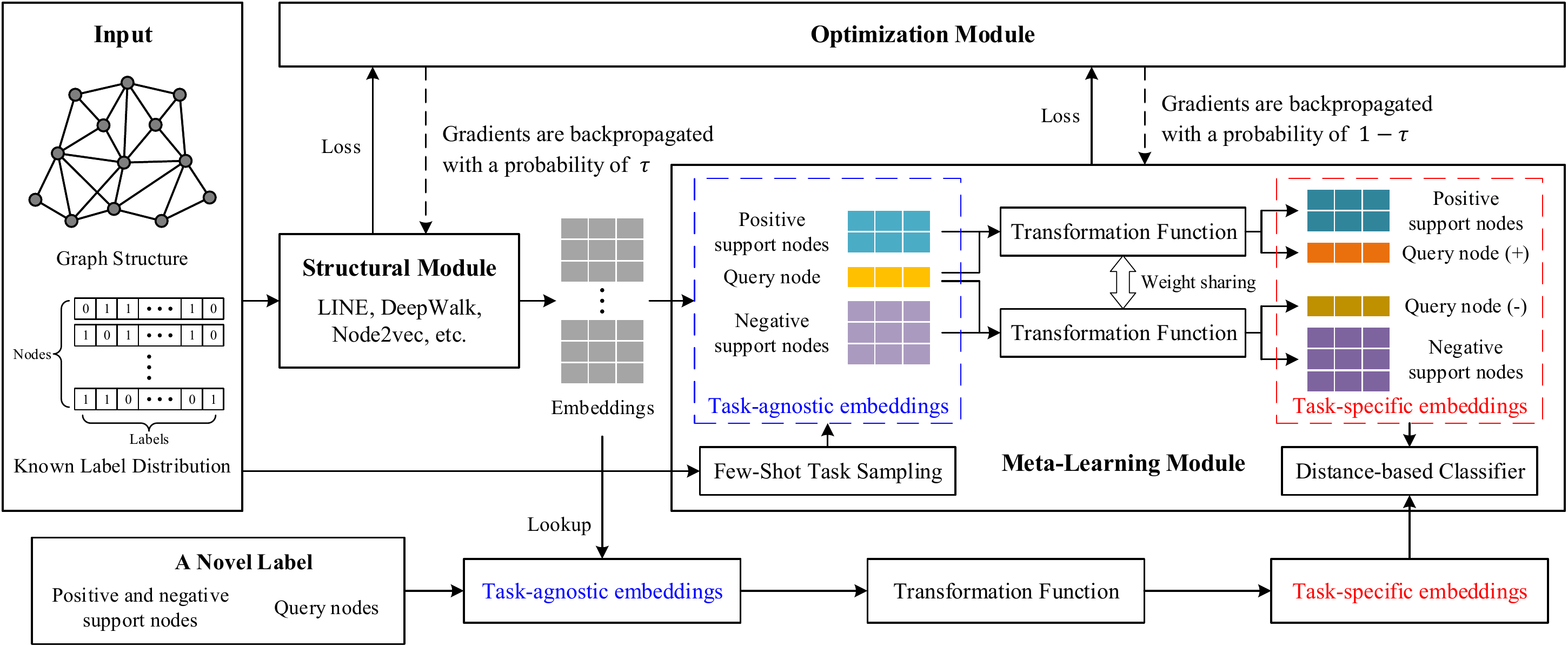}
\caption{
A schematic depiction of our \method{}.
In the meta-learning module, we use $2$ positive and $3$ negative support nodes for simplicity of illustration.
The threshold $\tau$ gradually decreases from $1$ to $0$ during training.
The flow of applying \method{} to a novel label is shown at the bottom.}
\label{fig:framework}
\end{figure*}

\section{Algorithm}

In this section, we present our proposed \method{} in detail,
which consists of three modules: the structural module, the meta-learning module, and the optimization module,
as shown in Fig.~\ref{fig:framework}.
Given a graph and some known labels, the structural module learns an embedding for each node based on the graph structure.
Then, the meta-learning module learns a transformation function that adapts the structure-only node embeddings for each few-shot node classification task sampled from the known labels and performs few-shot classification using a distance-based classifier.
Finally,
to optimize our model,
we propose a learning schedule that optimizes the structural and meta-learning modules with probabilities that gradually decrease and increase from $1$ to $0$ and from $0$ to $1$, respectively.

\subsection{Structural Module}
The structural module aims to learn a representation or embedding in the latent space for each node
while preserving the graph structure (i.e., the connections between nodes).
Mathematically, for each node $v_i\in\CV$,
we maximize the log-probability of observing its neighbors
by optimizing the following objective function:
$\min\sum_{v_i\in\CV}\sum_{v_j\in\CN(v_i)}\log\mathbb{P}(v_j|v_i)$,
where $\CN(v_i)$ denotes the neighboring nodes of $v_i$.
We optimize the above objective function following
the skip-gram architecture~\cite{mikolov2013distributed}.
Regarding the construction of the neighboring set $\CN(\cdot)$,
although there are many choices such as 1-hop neighbors based on the connectivity of nodes~\cite{tang2015line}
and the random walk strategy~\cite{perozzi2014deepwalk,grover2016node2vec},
in this paper we adopt the 1-hop neighbors for the sake of simplicity.
By optimizing the above objective,
we are able to obtain an embedding matrix $\M{U}\in\mathbb{R}^{|V|\times d}$,
of which the $i$-th row $\V{u}_i$ indicates the representation of $v_i$.

\subsection{Meta-Learning Module}
As alluded before,
we cast the problem of \problem{} into the meta-learning framework~\cite{ravi2016optimization,finn2017model} and simulate the few-shot setting with $\CY_{\text{known}}$ during training.
In what follows,
we first describe how to organize the graph structure and the known labels in the meta-learning scenario.
Then, we give a metric-based meta-learning paradigm for solving \problem{}.
In particular,
we propose a transformation function that
transforms the task-agnostic embeddings to the task-specific ones
in order to better deal
with the multi-label setting where each node may be associated with multiple labels.

\subsubsection{\textbf{Data Organization}}
\label{sec:data_org}
Instead of directly optimizing over the entire set of known labels
like traditional semi-supervised learning methods~\cite{yang2016revisiting},
we propose to construct a pool of few-shot node classification tasks according to the known labels $\CY_{\text{known}}$.
Analogous to few-shot image classification tasks
in the literature of meta-learning~\cite{ravi2016optimization},
a few-shot node classification task $\CT_i=(\CS_i,\CQ_i,y_i)$ is composed of
a support set $\CS_i$, a query set $\CQ_i$, and a label identifier $y_i$ randomly sampled from $\CY_{\text{known}}$.
The support set $\CS_i=\CS_{i}^{+}\cup\CS_{i}^{-}$
contains the set $\CS_{i}^{+}$ of randomly sampled positive nodes and the set $\CS_{i}^{-}$ of randomly sampled negative nodes,
where
$\CS_{i}^{+}\subset\CD_{y_i}^{+}$
and
$\CS_{i}^{-}\subset\CD_{y_i}^{-}$.
The query set $\CQ_i=\CQ_{i}^{+}\cup\CQ_{i}^{-}$ is defined in the same way
but does not intersect with the support set, namely
$\CQ_{i}^{+}\subset\CD_{y_i}^{+}\setminus\CS_{i}^{+}$ and $\CQ_{i}^{-}\subset\CD_{y_i}^{-}\setminus\CS_{i}^{-}$.
The task is,
given the support set of node-label pairs,
finding a classifier $f_{\CT_i}$ which is able to predict the probability $\hat{\SL}_{v_q,y_i}\in[0,1]$ for each query node $v_q$
with a low misclassification rate.
We denote by $\CT_i\sim p(\CT|\CY_{\text{known}})$ sampling a few-shot node classification task from $\CY_{\text{known}}$.

\subsubsection{\textbf{Meta-Learning with Embedding Transformation for \problem{}}}
\label{sec:simple_meta}
To facilitate learning to classify for a label with few associated nodes in a graph,
we apply a meta-learning flavored learning scheme.
Following the above definition of few-shot node classification tasks,
for each task $\CT_i=(\CS_i,\CQ_i,y_i)\sim p(\CT|\CY_{\text{known}})$,
we aim to construct a classifier $f_{\CT_i}$ for the label $y_i$ given the support set $\CS_i$,
which is able to classify the query nodes in the set $\CQ_i$.
Formally, for each $(v_q,\SL_{v_q,y_i})\in\CQ_i$,
the classification loss is defined as follows:
\begin{equation}
\CL(\hat{\SL}_{v_q,y_i},\SL_{v_q,y_i})=
-\SL_{v_q,y_i}\log\hat{\SL}_{v_q,y_i}
-(1-\SL_{v_q,y_i})\log(1-\hat{\SL}_{v_q,y_i}),
\end{equation}
where $\hat{\SL}_{v_q,y_i}$ denotes the predicted probability that
$v_q$ holds label $y_i$.
Here, to calculate the probability,
we adopt a distance-based classifier
which is commonly used in the metric-based meta-learning literature~\cite{snell2017prototypical}.
Specifically, for each task $\CT_i$,
the classifier $f_{\CT_i}$ is parametrized by two $d$-dimensional latent representations,
$\V{c}_+^{(i)} $ (called positive prototype) and $\V{c}_-^{(i)}$ (called negative prototype),
that correspond to the cases of holding and not holding label $y_i$, respectively. The predictions are made based on the distances between the node representations
and these two prototypes.
Mathematically, given the embedding vector $\V{u}_q$ of each query node $v_q$,
we have the predicted probability as
\begin{equation}
\label{eqn:prob_l}
\hat{\SL}_{v_q,y_i}
=f_{\CT_i}(v_q|\V{c}_+^{(i)},\V{c}_-^{(i)})
=\frac{\exp(-\dist{\V{u}_q}{\V{c}_+^{(i)}})}
{\sum_{m\in\{+,-\}}\exp(-\dist{\V{u}_q}{\V{c}_m^{(i)}})},
\end{equation}
where $\dist{\cdot}{\cdot}:\mathbb{R}^d\times\mathbb{R}^d\to\interval[open right,soft open fences]{0}{+\infty}$
is the squared Euclidean distance function
and the positive or negative prototype is usually calculated as the mean vector of node representations in the corresponding support set~\cite{snell2017prototypical}.

\noindent
\textbf{Why do we need Embedding Transformation?}
Equation~(\ref{eqn:prob_l}) makes predictions under the condition that
each node is represented by the same or \emph{task-agnostic} embedding vector
regardless of which label or task we are concerned about.
Technically,
this scheme makes sense for few-shot image classification in prior works~\cite{snell2017prototypical}
where each image is assigned to the same one and only one label.
However, this is problematic in the multi-label scenario
where each node could be assigned to multiple labels.
Here is an illustrating example.
In social networks,
suppose we have two classification tasks $\CT_1$ and $\CT_2$
with respect to different labels, namely ``Sports'' from $\CY_{\text{known}}$ and ``Music'' from $\CY_{\text{novel}}$,
and two users $A$ and $B$ are involved in these two tasks.
Both users $A$ and $B$ could give positive feedback to ``Sports'',
while on the other hand, they could give positive and negative feedback to ``Music'' respectively.
Intuitively, the task-agnostic scheme may provide similar embeddings after fitting well on the task $\CT_1$, which is not appropriate for the task $\CT_2$.

\noindent
\textbf{High-Level Module Design.}
To mitigate the above problem,
we propose to learn a
transformation function $Tr(\cdot)$
which transforms the task-agnostic embeddings
to some task-specific ones for each task.
First,
we argue that different query nodes have different correlation patterns with the nodes in the support set.
To fully explore how a query node correlates with the support nodes,
we propose to tailor the embeddings of the support nodes for each query node.
Second, to classify a query node, we are more interested in characterizing
the distance relationship between the query node and either positive or negative support nodes
rather than the relationship between the positive and negative support nodes.
Thus, during the transformation, we propose to adapt the query node with
the positive and the negative nodes in the support set separately.

Based on the above two principles, for each query node,
we first construct two sets:
one containing the task-agnostic embeddings of the query node and the positive support nodes,
and the other containing the task-agnostic embeddings of the query node and the negative support nodes.
Then, we separately feed the two sets into the transformation function.
The meta-learning module in Fig.~\ref{fig:framework} illustrates this process.
Formally,
given a task $\CT_i=(\CS_i,\CQ_i,y_i)$,
for each query node $v_q\in\CV_{\CQ_i}$,
we have
\begin{equation}
\label{eqn:adapt_sep}
\begin{split}
&\{\tilde{\V{u}}_{q,+}^{(i)}\}\cup\{\tilde{\V{u}}_{k,q}^{(i)}|v_k\in\CV_{\CS_i^{+}}\}
=Tr(\{\V{u}_q\}\cup\{\V{u}_k|v_k\in\CV_{\CS_i^{+}}\}), \\ %
&\{\tilde{\V{u}}_{q,-}^{(i)}\}\cup\{\tilde{\V{u}}_{k,q}^{(i)}|v_k\in\CV_{\CS_i^{-}}\}
=Tr(\{\V{u}_q\}\cup\{\V{u}_k|v_k\in\CV_{\CS_i^{-}}\}),
\end{split}
\end{equation}
where $\tilde{\V{u}}_{q,+}^{(i)}$
and $\tilde{\V{u}}_{q,-}^{(i)}$
denote the adapted embedding of the query node $v_q$
in relation to the positive and negative support nodes, respectively,
and $\tilde{\V{u}}_{k,q}^{(i)}$ denotes the adapted embedding of the support node $v_k$
tailored for the query node $v_q$.
As a result, each query node has
two different adapted embeddings $\tilde{\V{u}}_{q,+}^{(i)}$ and $\tilde{\V{u}}_{q,-}^{(i)}$
that are further used for comparisons with the adapted embeddings of the positive and negative support nodes, respectively.
A consequential benefit is that
the transformation function is more flexible to capture the multifaceted relationships between nodes in the multi-label scenario.
Imagine that even if the task-specific embeddings of the positive and negative support nodes or prototypes
are distributed close,
we are still able to make right predictions through altering
$\tilde{\V{u}}_{q,+}^{(i)}$ and $\tilde{\V{u}}_{q,-}^{(i)}$.
The ablation study in Section Experiments and the visualization in the supplement
confirm the superiority of this design.

\noindent
\textbf{Instantiation.}
As per the above discussions,
we propose to implement $Tr(\cdot)$ using
the self-attention architecture with the scaled dot-product attention mechanism~\cite{vaswani2017attention},
which has exhibited the ability to effectively capture relationships
between a set of elements.
We start with some basic concepts of the self-attention.
Each input element plays three different roles in the self-attention:
(1) It is compared with every other element to compute the weights that indicate \emph{how much it attends to other elements};
(2) It is compared with every other element to compute the weights that indicate \emph{how much other elements attend to it};
(3) It is used \emph{as part of the output of each element} after the weights between elements have been determined.
Following the prior work~\cite{vaswani2017attention},
for each input element,
we establish three vectors, the \emph{query} vector, the \emph{key} vector, and the \emph{value} vector,
to represent the three roles, respectively.
Typically, these three vectors are obtained by applying linear transformations to the input vector of each element
with three trainable matrices,
which enables us to learn to make each element suit the three roles it needs to play.

Next, we elaborate on how to leverage the self-attention architecture to instantiate Eqn.~(\ref{eqn:adapt_sep}),
which separately takes as input the two sets
$\{\V{u}_q\}\cup\{\V{u}_k|v_k\in\CV_{\CS_i^m}\}$ where $m\in\{+,-\}$.
For any two nodes $v_i,v_j\in\{v_q\}\cup\CV_{\CS_i^m}$ ($v_i$ and $v_j$ could be the same),
we first calculate the attention $\omega_{ij}$ that $v_i$ pays to $v_j$
as follows:
\begin{equation}\label{eqn:self-attention-weight}
\omega_{ij}=\frac
{\exp((\M{W}_Q\V{u}_i)\dotproduct(\M{W}_K\V{u}_j)/{d^{\prime}}^{1/2})}
{\sum_{v_k\in\{v_q\}\cup\CV_{\CS_i^m}}
\exp((\M{W}_Q\V{u}_i)\dotproduct(\M{W}_K\V{u}_k)/{d^{\prime}}^{1/2})},
\end{equation}
where $\M{W}_Q,\M{W}_K\in\mathbb{R}^{d^{\prime}\times d}$ denote the trainable matrices
that project the input vectors into the \emph{query} and \emph{key} vectors, respectively,
$d^{\prime}$ denotes the dimension of the \emph{query}, \emph{key}, and \emph{value} vectors,
``$\dotproduct$'' denotes the dot product operator,
and $\frac{1}{\sqrt{d^{\prime}}}$ is a scaling factor to avoid extremely small gradients~\cite{vaswani2017attention}.
In effect, the attention $\omega_{ij}$ reflects the degree to which the node $v_j$ relates to or influences $v_i$.
Then, the output or transformed vector of each node aggregates information from every other node in a weighted manner.
Specifically,
let $\M{W}_V\in\mathbb{R}^{d^{\prime}\times d}$ be the trainable matrix
to calculate the \emph{value} vectors
and $\M{W}_O\in\mathbb{R}^{d\times d^{\prime}}$ be another trainable matrix
to ensure that the output vectors are of the same dimension as the input vectors.
We compute the output vector of the query node $v_q$ as
\begin{equation}
\label{eqn:self-attention-query}
\tilde{\V{u}}_{q,m}^{(i)}=
\M{W}_O\biggl(
\omega_{qq}\M{W}_V\V{u}_q+\sum_{v_k\in\CV_{\CS_i^m}}
\omega_{qk}\M{W}_V\V{u}_k\biggr),
\end{equation}
and compute the output vector of each support node $v_k\in\CV_{\CS_i^m}$
tailored for the query node $v_q$ as
\begin{equation}
\label{eqn:self-attention-support}
\tilde{\V{u}}_{k,q}^{(i)}=
\M{W}_O\biggl(
\omega_{kk}\M{W}_V\V{u}_k+\sum_{v_j\in\bigl(\CV_{\CS_i^m}\setminus\{v_k\}\bigr)\cup\{v_q\}}
\omega_{kj}\M{W}_V\V{u}_j\biggr).
\end{equation}
We refer readers to the supplement for more details
on the instantiation of the transformation function.

With the transformed embeddings, we further
calculate the positive and negative prototypes tailored for $v_q$
as well as the predicted probability
as follows:
\begin{equation}
\label{eqn:new_prob_l}
\tilde{\V{c}}_{m,q}^{(i)}=\frac{1}{|\CS_{i}^{m}|}\sum_{v_k\in\CV_{\CS_i^{m}}}\tilde{\V{u}}_{k,q}^{(i)},
\; m\in\{+,-\},
\; \text{and} \;
\hat{\SL}_{v_q,y_i}
=\frac{\exp(-\dist{\tilde{\V{u}}_{q,+}^{(i)}}{\tilde{\V{c}}_{+,q}^{(i)}})}
{\sum_{m\in\{+,-\}}\exp(-\dist{\tilde{\V{u}}_{q,m}^{(i)}}{\tilde{\V{c}}_{m,q}^{(i)}})}.
\end{equation}
The final meta-learning objective is formulated as:
\begin{equation}
\label{obj:meta_adapt}
\min_{\M{U},\Theta}\sum_{\CT_i}
\sum_{(v_q,\SL_{v_q,y_i})\in\CQ_i}
\CL(\hat{\SL}_{v_q,y_i},\SL_{v_q,y_i})
+\lambda\sum\|\Theta\|_2^2,
\end{equation}
where $\CT_i\sim p(\CT|\CY_{\text{known}})$, $\hat{\SL}_{v_q,y_i}$ is calculated through Eqn.~(\ref{eqn:new_prob_l}),
$\Theta$ refers to the set of parameter matrices
(e.g., $\M{W}_Q$, $\M{W}_K$, and $\M{W}_V$) contained in $Tr(\cdot)$,
and $\lambda>0$ is a balancing factor.

\subsection{Optimization and Using the Learned Model for Few-Shot Novel Labels}

For optimization,
one typical way is to minimize the (weighted) sum of the structural loss and the meta loss.
However,
the structure information of the graph is still not properly embedded at the beginning of the training stage,
and the node representations are somewhat random which make no sense for the few-shot classification tasks.
Therefore, a training procedure that focuses on optimizing the structural module
at the beginning and then gradually pays more attention to optimizing the meta-learning module is preferably required.
To satisfy this requirement,
we take inspiration from learning rate annealing~\cite{krizhevsky2012imagenet} and introduce a probability threshold $\tau$,
and in each training step the structural and meta modules are optimized
with probabilities of $\tau$ and $1-\tau$, respectively.
The probability threshold $\tau$ is gradually decayed from $1$ to $0$
in a staircase manner,
namely $\tau=1/(1+\gamma\left\lfloor\frac{step}{N_{\text{decay}}}\right\rfloor)$
where $\gamma$ is the decay rate, $step$ is the current step number,
and $N_{\text{decay}}$ indicates how often the threshold is decayed.
The complete optimization procedure is outlined in 
the supplement.
In addition, the time complexity is analyzed in 
the supplement.

Recall that our ultimate goal is to,
after observing a few support nodes associated with a novel label $y\in\CY_{\text{novel}}$,
predict whether other (or some query) nodes have the label $y$ or not.
In effect, this can be regarded as a few-shot node classification task
$\CT=(\CS,\CQ,y)$.
\begin{wraptable}{r}{0.4\linewidth}
\centering
\caption{Statistics of the datasets.}
\label{tab:dataset}
\resizebox*{\linewidth}{!}{%
\begin{tabular}{@{}lrrr@{}}
\toprule
Dataset & $\#$Nodes & $\#$Edges & $\#$Labels \\ \midrule
BlogCatalog & 10,312 & 333,983 & 39 \\
Flickr & 80,513 & 5,899,882 & 195 \\
PPI & 3,890 & 76,584 & 50 \\
Mashup & 16,143 & 300,181 & 28 \\ \bottomrule
\end{tabular}}
\end{wraptable}

After optimization, we have obtained the task-agnostic node representations $\M{U}$,
and the transformation function $Tr(\cdot)$ parameterized by $\Theta$.
Thus, to classify a query node $v_q\in\CQ$,
we simply look up the representations of the query node and the support nodes from $\M{U}$,
adapt their representations using the transformation function as formulated in Eqn.~(\ref{eqn:self-attention-query})~and~(\ref{eqn:self-attention-support}),
and compute the predicted probability according to Eqn.~(\ref{eqn:new_prob_l}).
The detailed procedure is presented in 
the supplement.

\section{Experiments}\label{sec:expt}

Four publicly available real-world benchmark datasets
are used to validate the effectiveness of our method.
The statistics of these datasets are summarized in Table~\ref{tab:dataset}.
For each dataset, we split the labels into training, validation, and test labels according to a ratio of 6:2:2. In the training stage, we regard the training labels as the known labels and sample few-shot node classification tasks from them.
For validation and test purposes,
we regard the validation and test labels as the novel labels
and sample 1,000 tasks from them, respectively.
We use the average classification performance on the test tasks for comparisons of different methods.
For ease of presentation,
we use
$K_{\CS,+}$, $K_{\CS,-}$, $K_{\CQ,+}$, and $K_{\CQ,-}$
to indicate the respective numbers of positive support, negative support,
positive query, and negative query nodes in a task.
We compare \method{} with Label Propagation~\cite{zhu2003semi},
unsupervised network embedding methods
(LINE~\cite{tang2015line} and Node2vec~\cite{grover2016node2vec}),
semi-supervised network embedding methods
(Planetoid~\cite{yang2016revisiting} and GCN~\cite{kipf2016semi}),
and Meta-GNN~\cite{zhou2019meta}.
For detailed experimental settings including dataset and baseline descriptions, baseline evaluation procedure, and parameter settings,
please refer to
the supplement.

\begin{table*}[t]
\centering
\caption{Results on few-shot node classification tasks with novel labels. OOM means out of memory (16 GB GPU memory).
The standard deviation is provided in the supplement.}
\label{tab:expt_1}
\subcaptionbox{$K_{\ast,+}=10$ and $K_{\ast,-}=20$.}{
\resizebox{\linewidth}{!}{%
\begin{tabular}{@{}lrrrrrrrrrrrr@{}}
\toprule
\multirow{2}{*}{Method} & \multicolumn{3}{c}{BlogCatalog} & \multicolumn{3}{c}{Flickr} & \multicolumn{3}{c}{PPI} & \multicolumn{3}{c}{Mashup} \\ \cmidrule(lr){2-4} \cmidrule(lr){5-7} \cmidrule(lr){8-10} \cmidrule(lr){11-13}
 & \multicolumn{1}{c}{AUC} & \multicolumn{1}{c}{F$_1$} & \multicolumn{1}{c}{Recall} & \multicolumn{1}{c}{AUC} & \multicolumn{1}{c}{F$_1$} & \multicolumn{1}{c}{Recall} & \multicolumn{1}{c}{AUC} & \multicolumn{1}{c}{F$_1$} & \multicolumn{1}{c}{Recall} & \multicolumn{1}{c}{AUC} & \multicolumn{1}{c}{F$_1$} & \multicolumn{1}{c}{Recall} \\ \midrule
LP & 0.6422 & 0.1798 & 0.2630 & 0.8196 & 0.4321 & 0.4989 & 0.6285 & 0.2147 & 0.2769 & 0.6488 & 0.3103 & 0.4535 \\
LINE & 0.6690 & 0.2334 & 0.1595 & 0.8593 & 0.6194 & 0.5418 & 0.6372 & 0.2147 & 0.1456 & 0.6926 & 0.2970 & 0.2142 \\
Node2vec & 0.6697 & 0.3750 & 0.2940 & 0.8504 & 0.6664 & 0.6147 & 0.6273 & 0.3545 & 0.2860 & 0.6575 & 0.3835 & 0.3147 \\
Planetoid & 0.6850 & 0.4657 & 0.4301 & \textbf{0.8601} & 0.6638 & 0.6331 & 0.6791 & 0.4672 & 0.4411 & 0.7056 & 0.4825 & 0.4218 \\
GCN & 0.6643 & 0.3892 & 0.3379 & OOM & OOM & OOM & 0.6596 & 0.4176 & 0.3729 & 0.6910 & 0.4065 & 0.3607 \\
Meta-GNN & 0.6533 & 0.3567 & 0.2962 & OOM & OOM & OOM & 0.6537 & 0.3964 & 0.3373 & 0.7093 & 0.4689 & 0.4202 \\ \midrule
\method{} & \textbf{0.6986} & \textbf{0.5380} & \textbf{0.6203} & 0.8462 & \textbf{0.7118} & \textbf{0.7700} & \textbf{0.6865} & \textbf{0.5188} & \textbf{0.5621} & \textbf{0.7645} & \textbf{0.5764} & \textbf{0.5566} \\
\%Improv. & 1.99 & 15.53 & 44.22 & -1.62 & 6.81 & 21.62 & 1.09 & 11.04 & 27.43 & 7.78 & 19.46 & 22.73 \\ \bottomrule
\end{tabular}%
}
}\\
\vspace{10pt}
\subcaptionbox{$K_{\ast,+}=10$ and $K_{\ast,-}=40$.}{
\resizebox{\linewidth}{!}{%
\begin{tabular}{@{}lrrrrrrrrrrrr@{}}
\toprule
\multirow{2}{*}{Method} & \multicolumn{3}{c}{BlogCatalog} & \multicolumn{3}{c}{Flickr} & \multicolumn{3}{c}{PPI} & \multicolumn{3}{c}{Mashup} \\ \cmidrule(lr){2-4} \cmidrule(lr){5-7} \cmidrule(lr){8-10} \cmidrule(lr){11-13}
 & \multicolumn{1}{c}{AUC} & \multicolumn{1}{c}{F$_1$} & \multicolumn{1}{c}{Recall} & \multicolumn{1}{c}{AUC} & \multicolumn{1}{c}{F$_1$} & \multicolumn{1}{c}{Recall} & \multicolumn{1}{c}{AUC} & \multicolumn{1}{c}{F$_1$} & \multicolumn{1}{c}{Recall} & \multicolumn{1}{c}{AUC} & \multicolumn{1}{c}{F$_1$} & \multicolumn{1}{c}{Recall} \\ \midrule
LP & 0.6421 & 0.0554 & 0.0727 & 0.8253 & 0.3055 & 0.3040 & 0.6298 & 0.0773 & 0.0748 & 0.6534 & 0.1156 & 0.1284 \\
LINE & 0.6793 & 0.0529 & 0.0328 & 0.8644 & 0.4154 & 0.3485 & 0.6423 & 0.0496 & 0.0300 & 0.7009 & 0.0956 & 0.0617 \\
Node2vec & 0.6792 & 0.1982 & 0.1340 & 0.8558 & 0.5295 & 0.4602 & 0.6309 & 0.1894 & 0.1306 & 0.6643 & 0.2070 & 0.1447 \\
Planetoid & 0.6981 & 0.2980 & 0.2319 & \textbf{0.8728} & 0.5040 & 0.4461 & 0.6879 & 0.3100 & 0.2523 & 0.7095 & 0.3279 & 0.2551 \\
GCN & 0.6794 & 0.2104 & 0.1583 & OOM & OOM & OOM & 0.6608 & 0.2531 & 0.1974 & 0.7007 & 0.2558 & 0.2098 \\
Meta-GNN & 0.6724 & 0.2152 & 0.1618 & OOM & OOM & OOM & 0.6617 & 0.2575 & 0.2088 & 0.7140 & 0.3412 & 0.2864 \\ \midrule
\method{} & \textbf{0.7139} & \textbf{0.4398} & \textbf{0.5819} & 0.8505 & \textbf{0.6220} & \textbf{0.7460} & \textbf{0.7039} & \textbf{0.4298} & \textbf{0.5327} & \textbf{0.7684} & \textbf{0.4814} & \textbf{0.4816} \\
\%Improv. & 2.26 & 47.58 & 150.93 & -2.55 & 17.47 & 62.10 & 2.33 & 38.65 & 111.14 & 7.62 & 41.09 & 68.16 \\ \bottomrule
\end{tabular}%
}
}
\end{table*}

\noindent
\textbf{Overall Comparisons.}
Following the standard evaluation protocol of meta-learning~\cite{finn2017model},
we first compare different methods
with $K_{\CS,+}=K_{\CQ,+}$ and $K_{\CS,-}=K_{\CQ,-}$
(hereafter using $K_{\ast,+}$ and $K_{\ast,-}$ for simplicity),
and these numbers are the same for both training and test tasks.
Considering that negative samples are usually easier than positive samples to acquire
we report the overall performance
with $K_{\ast,+}$
set to 10 and $K_{\ast,-}$ set to 20 and 40, respectively.
The comparison results on the four datasets
are presented in Table~\ref{tab:expt_1}.
Since in our application scenarios
we prefer to discover proteins with new functions in biological networks
and find users who are interested in the latest advertisements on online social networks
rather than predict negative samples accurately,
we report Recall in addition to AUC and F$_1$.
To eliminate randomness, all of the results here and in the following quantitative experiments are averaged over 50 different trials.

From Table~\ref{tab:expt_1},
we observe that
\method{} consistently and significantly outperforms all other methods in terms of the three metrics
across all the four datasets except the AUC scores on Flickr dataset.
By jointly analyzing the F$_1$ and Recall scores,
\method{} predicts positive nodes from imbalanced data more effectively than the baselines, with little loss of precision.
In particular, \method{} achieves 44.22\% and 150.93\% gains over the strongest baseline (i.e., Planetoid)
with respect to Recall on BlogCatalog dataset when $K_{\ast,-}$ equals 20 and 40, respectively.

Compared with the unsupervised methods, Planetoid reaches better performance owing to
the use of training labels.
On the other hand, GCN also uses training labels as supervision, while does not show satisfactory performance and even worse performance than Node2vec,
which is due to that the graph convolution relies heavily on node attributes for feature propagation and aggregation as mentioned before
and the lack of node attributes limits its representativeness and thus classification capacity.

Besides, Meta-GNN underperforms the unsupervised methods and GCN in some cases,
which seems to contradict the published results in the original paper.
The reasons are twofold:
(1) Meta-GNN is built upon GCN and the predictive ability is also limited due to the lack of node attributes, while the original paper focuses on attributed graphs;
(2) Meta-GNN simply applies MAML to GCN
and is originally used for the multi-class setting
(e.g., each document has the same and only one label in Cora~\cite{sen2008collective}).
However, we consider the multi-label setting and the same pair of nodes may have opposite relations in different tasks, which will introduce noisy and contradictory signals
in the optimization process of MAML and further degrade the performance in some cases.

\begin{wraptable}{r}{0.5\linewidth}
\centering
\caption{Results of ablation study in terms of F$_1$.}
\label{tab:ablation}
\resizebox{\linewidth}{!}{%
\begin{tabular}{@{}lrrrr@{}}
\toprule
\multirow{2}{*}{Method} & \multicolumn{2}{c}{$K_{\ast,+}=10,K_{\ast,-}=20$} & \multicolumn{2}{c}{$K_{\ast,+}=10,K_{\ast,-}=40$} \\ \cmidrule(lr){2-3} \cmidrule(lr){4-5}
 & \multicolumn{1}{c}{BlogCatalog} & \multicolumn{1}{c}{PPI} & \multicolumn{1}{c}{BlogCatalog} & \multicolumn{1}{c}{PPI} \\ \midrule %
\method{} & \textbf{0.5380} & \textbf{0.5188} & \textbf{0.4398} & \textbf{0.4298} \\ \midrule
V1 & 0.5028 & 0.4851 & 0.3998 & 0.3721 \\
V2 & 0.5020 & 0.5011 & 0.4141 & 0.4078 \\
V3 & 0.5205 & 0.4980 & 0.4039 & 0.4074 \\
V4 & 0.4748 & 0.4614 & 0.3549 & 0.3389 \\
V5 & 0.4892 & 0.4819 & 0.3699 & 0.3777 \\ \bottomrule
\end{tabular}%
}
\end{wraptable}

\noindent
\textbf{Ablation Study.}
In what follows,
to gain deeper insight into the contributions of different components involved in our approach,
we conduct ablation studies by considering the following variants:
(1) a variant
without the transformation function;
(2) a variant that produces task-specific embeddings by simply feeding all support and query node representations into the self-attention network instead of
according to
Eqn.~(\ref{eqn:adapt_sep});
(3) a variant that optimizes the total loss of the two modules with the meta-learning loss scaled
by a balancing factor searched over $\{10^{-2},10^{-1},\cdots,10^{2}\}$;
(4) the node embeddings are learned at the beginning and then left fixed
(i.e., the structural and meta-learning losses are optimized separately);
(5) each node is represented by a one-hot vector and the node embeddings are only optimized with respect to the meta-learning loss.
We refer to these variants as V1, V2, V3, V4, and V5.
The results are summarized in Table~\ref{tab:ablation}.

We see that \method{} consistently outperforms its ablated variants.
Especially,
comparing \method{} with V1,
we confirm the necessity to introduce the transformation function.
The comparison with V2 demonstrates the effectiveness of
our special design in Eqn~(\ref{eqn:adapt_sep}).
The results of V3 and V4 indicate that our proposed scheduling strategy
can boost the performance of \method{}
with a better balance between the two modules during optimization.
The results of V5 show that it is important to introduce the structural loss to optimize the node embeddings.
In addition,
we see that
V1 underperforms V2 even if the node embeddings of V1 are first learned from the graph structure.
We speculate that the reason is that
at the beginning,
the latent space of node embeddings somewhat overfits to the metric of graph structure learning,
making it harder to adapt to the metric of subsequent meta-learning or few-shot learning tasks.

\noindent
\textbf{Additional Experiments.}
In the supplement,
we present more analytical experiments on the numbers of support and query nodes,
and illustrate the effect of the proposed transformation function
through a visualization experiment.

\section{Conclusion and Future Work}
This paper 
studies the problem of node classification on graphs
with few-shot novel labels.
To address this problem,
we propose a new semi-supervised framework \method{} that integrates network embedding and meta-learning.
Benefiting from utilizing known labels in a meta-learning manner,
\method{} is able to automatically capture the relationships between the graph structure and the node labels as prior knowledge
and make use of the prior knowledge to help recognize novel labels with only a few support nodes.
Extensive experiments on four real-world datasets
demonstrate the superiority of our proposed method.
In the future,
to improve the interpretability, we plan to extend our approach to quantify the relationships between different labels (e.g., the weight that one label contributes to another) during meta-learning.
Another interesting idea is to explicitly incorporate the graph structure information into the meta-learning module,
such as developing a more principled way to construct few-shot tasks according to the graph structure instead of random sampling.

\section*{Broader Impact}

In general, this work has potential positive impact on graph-related fields that need to deal with the classification problem with respect to few-shot novel labels.
For instance,
our work is beneficial for social networking service providers such as Facebook and Twitter.
These providers can obtain quick and effective feedback on newly developed features through distributing surveys among a small group of users on social networks.
In addition, our work can also help biologists, after discovering a new function of certain existing proteins,
quickly understand whether other proteins in a protein-protein interaction network have the new function,
which improves the efficiency of wet laboratory experimentation.
Moreover,
many recommender systems model users and items as a graph and enhance the recommendation performance with the aid of network embedding.
To some extent, our work is potentially useful to alleviate the cold-start problem as well.

At the same time,
our model could be biased towards the few-shot setting after training
and not provide superior performance on those labels with many support nodes.
In practice, if the original few-shot label gradually has enough support nodes
(e.g., biologists identify more proteins with and without the new function through laboratory experiments),
we recommend using general unsupervised or semi-supervised methods (e.g., Node2vec~\cite{grover2016node2vec} or Planetoid~\cite{yang2016revisiting}) to recognize the label.

\begin{ack}
The research presented in this paper is supported in part by National Natural Science Foundation of China (61922067, U1736205, 61902305), MoE-CMCC ``Artifical Intelligence'' Project (MCM20190701), Natural Science Basic Research Plan in Shaanxi Province of China (2019JM-159), Natural Science Basic Research Plan in Zhejiang Province of China (LGG18F020016).
\end{ack}

\bibliographystyle{plainnat}
\bibliography{references}

\newpage
\appendix

\section{Additional Algorithm Details}
\subsection{\textbf{Details of the Transformation Function}}\label{sec:details_trans}

\begin{wrapfigure}{r}{0.55\linewidth}
\centering
\includegraphics[width=\linewidth]{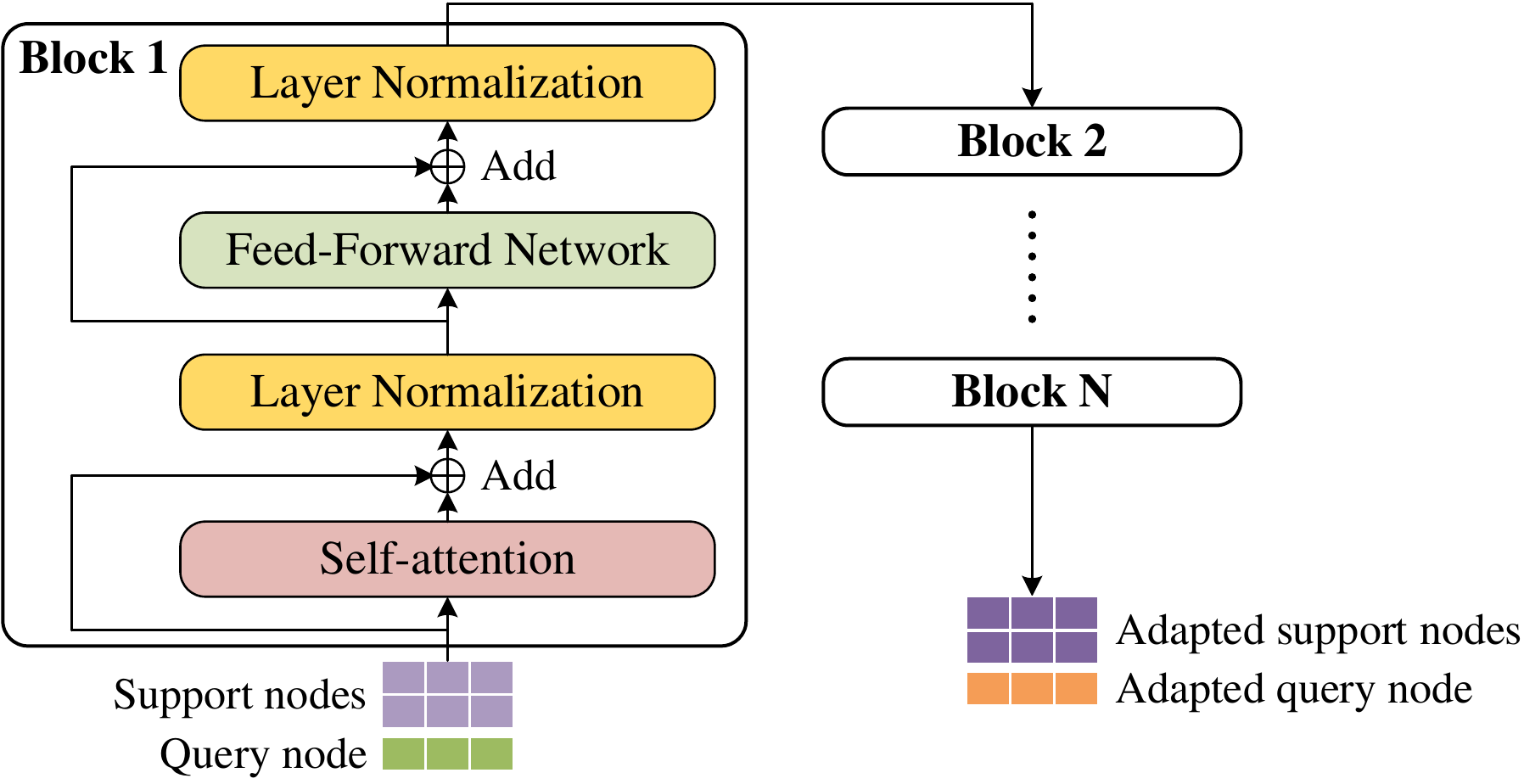}
\caption{Illustration of the transformation function.
The support nodes are either positive or negative.}
\label{fig:appen_framework}
\end{wrapfigure}

For the transformation function, we stack multiple computation blocks
as shown in Fig.~\ref{fig:appen_framework}.
The stacking mechanism helps the function capture comprehensive relationships between nodes
such that the performance is boosted.
In each computation block, there are mainly two modules.
The first is a self-attention module used to capture the relationships between input nodes,
and the second is a node-wise fully-connected feed-forward network used to introduce nonlinearity.
In addition, following~\cite{vaswani2017attention}, we employ a residual connection around each of the self-attention module and the feed-forward network and then perform layer normalization,
in order to make the optimization faster and more stable.

\begin{figure}[t]
\centering
\includegraphics[width=0.98\linewidth]{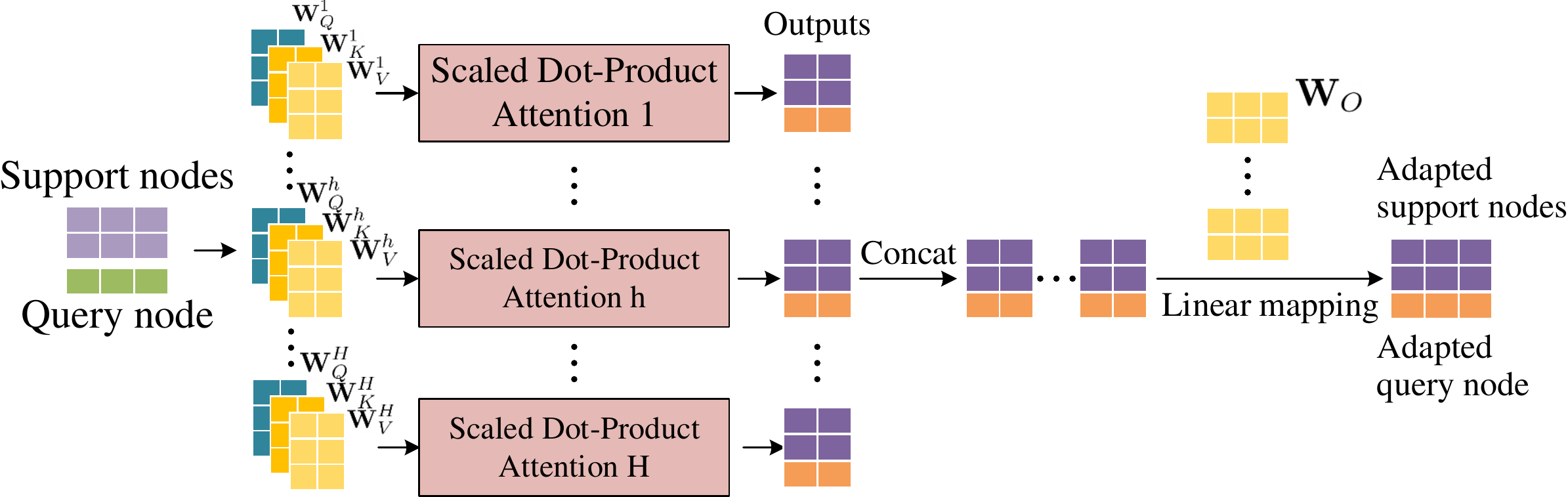}
\caption{Illustration of the self-attention module.
The support nodes are either positive or negative.}
\label{fig:appen_attention}
\end{figure}

The detailed architecture of the self-attention module is illustrated in Fig.~\ref{fig:appen_attention}. Following~\cite{vaswani2017attention}, we extend the self-attention
with multiple parallel attention heads using multiple sets of trainable matrices
(i.e., $\M{W}_Q^h, \M{W}_K^h, \M{W}_V^h \in \mathbb{R}^{\frac{d^{\prime}}{H}\times d}$ where $h=1,\ldots,H$).
In each attention head (i.e., each scaled dot-product attention),
for any two nodes $v_i,v_j\in\{v_q\}\cup\CV_{\CS_i^m}$ ($v_i$ and $v_j$ could be the same and $m\in\{+,-\}$) within task $\CT_i$,
we first calculate the attention $\omega_{ij}$ that $v_i$ pays to $v_j$ as follows:
\begin{equation}\label{eqnsupp:self-attention-weight}
\omega_{ij}^h=\frac
{\exp((\M{W}_Q^h\V{u}_i)\dotproduct(\M{W}_K^h\V{u}_j)/\sqrt{d^{\prime}/H})}
{\sum_{v_k\in\{v_q\}\cup\CV_{\CS_i^m}}
\exp((\M{W}_Q^h\V{u}_i)\dotproduct(\M{W}_K^h\V{u}_k)/\sqrt{d^{\prime}/H})},
\end{equation}
where ``$\dotproduct$'' denotes the dot product operator.
Then,
we compute the output vector of the query node $v_q$ as
\begin{equation}
\label{eqnsupp:self-attention-query}
\tilde{\V{u}}_{q,m}^{i,h}=
\omega_{qq}^h\M{W}_V^h\V{u}_q+\sum_{v_k\in\CV_{\CS_i^m}}
\omega_{qk}^h\M{W}_V^h\V{u}_k,
\end{equation}
and compute the output vector of each support node $v_k\in\CV_{\CS_i^m}$
tailored for the query node $v_q$ as
\begin{equation}
\label{eqnsupp:self-attention-support}
\tilde{\V{u}}_{k,q}^{i,h}=
\omega_{kk}^h\M{W}_V^h\V{u}_k+\sum_{v_j\in\bigl(\CV_{\CS_i^m}\setminus\{v_k\}\bigr)\cup\{v_q\}}
\omega_{kj}^h\M{W}_V^h\V{u}_j.
\end{equation}
Finally,
we concatenate the output vectors of all attention heads
and use a trainable matrix $\M{W}_O\in\mathbb{R}^{d\times d^{\prime}}$ to project the concatenated vectors
into the original space with the input dimension:
\begin{equation}
\label{eqnsupp:self-attention-final}
\tilde{\V{u}}_{q,m}^{(i)} = \M{W}_O(\tilde{\V{u}}_{q,m}^{i,1} \oplus \cdots \oplus \tilde{\V{u}}_{q,m}^{i,H}),
\quad \text{and} \quad
\tilde{\V{u}}_{k,q}^{(i)} = \M{W}_O(\tilde{\V{u}}_{k,q}^{i,1} \oplus \cdots \oplus \tilde{\V{u}}_{k,q}^{i,H}),
\forall v_k\in\CV_{\CS_i^m}.
\end{equation}

The multiple parallel attention heads
allow the function to jointly attend to information from different input nodes for each input node,
and thus help the function better exploit the relationships between input nodes.

\subsection{Pseudo Codes}
The optimization procedure is outlined in Algorithm~\ref{alg:procedure}.
The procedure of using the learned model for few-shot novel labels is presented in Algorithm~\ref{alg:procedure_adapt}.

\subsection{\textbf{Time Complexity Analysis}}\label{sec:time}
For the structural module, we optimize the objective function
in a way similar to~\cite{tang2015line} and the time complexity
is $O(kd|\CE|)$ where $k$ is the number of negative nodes at each iteration,
$d$ is the dimension of node embeddings, and $|\CE|$ is the number of edges.
For the meta-learning module,
the time cost mainly comes from the embedding transformation through the self-attention architecture~\cite{vaswani2017attention}.
Specifically,
let $m$ be the number of query nodes and $n$ be the number of positive or negative support nodes.
Calculating the \emph{query}, \emph{key}, and \emph{value} vectors
takes $O(mndd^{\prime})$,
where $d^{\prime}$ is the dimension of the \emph{query}, \emph{key}, and \emph{value} vectors.
Calculating the attention weights and the weighted sum of \emph{value} vectors
takes $O(mn^2d^{\prime})$.
Calculating the final output vectors takes $O(mndd^{\prime})$.
Overall, the time complexity of \method{}
is $O(kd|\CE| + mndd^{\prime} + mn^2d^{\prime})$.
Note that we can take advantage of GPU acceleration for optimization in practice.

\section{Details of the Experimental Settings}\label{sec:details_expt}

\subsection{Datasets}

Four datasets are used in our experiments.

\noindent
\textbf{BlogCatalog}~\cite{tang2009relational}: This dataset is the friendship network crawled from the BlogCatalog website.
The friendships and group memberships are encoded in the edges and labels, respectively.\ignorespaces
\footnote{\url{http://socialcomputing.asu.edu/datasets/BlogCatalog3}}

\noindent
\textbf{Flickr}~\cite{tang2009relational}: This dataset is the friendship network among the bloggers crawled from the Flickr website. The friendships and group memberships are encoded in the edges and the labels, respectively.\ignorespaces
\footnote{\url{http://socialcomputing.asu.edu/datasets/Flickr}}

\noindent
\textbf{PPI}~\cite{grover2016node2vec}:
This dataset is a protein-protein interaction network for Homo Sapiens.
Different labels represent different function annotations of proteins.\ignorespaces
\footnote{\url{https://snap.stanford.edu/node2vec/}}

\noindent
\textbf{Mashup}~\cite{yue2019graph}:
This dataset is a protein-protein interaction network for human.
Different labels represent different function annotations of proteins.\ignorespaces
\footnote{\url{https://github.com/xiangyue9607/BioNEV}}

\begin{algorithm}[!t]
\caption{The Optimization Procedure of \method{}}
\label{alg:procedure}
\begin{algorithmic}[1]
\REQUIRE
Graph $G$, total number of steps $N$, decay rate $\gamma$, decay period $N_{\text{decay}}$
\ENSURE
The embedding matrix $\M{U}\in\mathbb{R}^{|V|\times M}$,
the function $Tr(\cdot)$
\STATE Randomly initialize $\M{U}$
and the parameters $\Theta$ of $Tr(\cdot)$
\FOR{$step=0$ \TO $N$}
\STATE Calculate the threshold
$\tau=1/(1+\gamma\left\lfloor\frac{step}{N_{\text{decay}}}\right\rfloor)$
\STATE Draw a random number $r\sim \text{Uniform}(0,1)$
\IF[Optimize the structural module]{$r<\tau$}
\STATE Sample a batch of pairs $\{(v_i,v_j)|v_i\in\CV,v_j\in\CN(v_i)\}$
\STATE Update $\M{U}$
to optimize the objective function: \\
\begin{equation}
\min\sum_{v_i\in\CV}\sum_{v_j\in\CN(v_i)}\log\mathbb{P}(v_j|v_i)
\end{equation}
\ELSE[Optimize the meta-learning module]
\STATE Sample a batch of tasks $\CT_i$ from $\CY_{known}$
\FORALL{$\CT_i=(\CS_i,\CQ_i,y_i)$}
\FORALL{$v_q\in\CQ_i$}
\STATE Calculate the adapted embeddings
$\{\tilde{\V{u}}_{k,q}^{(i)}|v_k\in\CV_{\CS_i^m}\}$ and $\tilde{\V{u}}_{q,m}^{(i)}$,
where $m\in\{+,-\}$,
via Eqn.~(\ref{eqnsupp:self-attention-final})
\STATE Calculate the prototypes $\tilde{\V{c}}_{+,q}^{(i)}$ and $\tilde{\V{c}}_{-,q}^{(i)}$: \\
\begin{equation}\label{eqn:protos}
\tilde{\V{c}}_{m,q}^{(i)}=\frac{1}{|\CS_{i}^{m}|}\sum_{v_k\in\CV_{\CS_i^{m}}}\tilde{\V{u}}_{k,q}^{(i)},
\; m\in\{+,-\}
\end{equation}
\STATE Calculate the predicted probability that $v_q$ holds $y_i$:
\begin{equation}\label{eqn:prob}
\hat{\SL}_{v_q,y_i}
=\frac{\exp(-\dist{\tilde{\V{u}}_{q,+}^{(i)}}{\tilde{\V{c}}_{+,q}^{(i)}})}
{\sum_{m\in\{+,-\}}\exp(-\dist{\tilde{\V{u}}_{q,m}^{(i)}}{\tilde{\V{c}}_{m,q}^{(i)}})}
\end{equation}
\ENDFOR
\ENDFOR
\STATE Update $\M{U}$ and $\Theta$ to optimize the objective function:
\begin{equation}
\min_{\M{U},\Theta}\sum_{\CT_i}
\sum_{(v_q,\SL_{v_q,y_i})\in\CQ_i}
\CL(\hat{\SL}_{v_q,y_i},\SL_{v_q,y_i})
+\lambda\sum\|\Theta\|_2^2,
\end{equation}
\ENDIF
\ENDFOR
\end{algorithmic}
\end{algorithm}

\begin{algorithm}[!t]
\caption{Applying \method{} to Few-Shot Novel Labels}
\label{alg:procedure_adapt}
\begin{algorithmic}[1]
\REQUIRE
The embedding matrix $\M{U}$,
the function $Tr(\cdot)$,
a novel label $y\in\CY_{novel}$,
associated positive support nodes $\CV_{\CS^{+}}$ and negative support nodes $\CV_{\CS^{-}}$,
query nodes $\CV_{\CQ}$
\ENSURE%
The predicted probability $\hat{\SL}_{v_q,y}$ for each query node $v_q$
\STATE Look up in $\M{U}$ to get the support and query embeddings $\mathbf{u}_{k}, \mathbf{u}_{q}$.
\FOR{$v_q$ in $\CV_{\CQ}$}
    \STATE Adapt $v_q$ together with $\CV_{\CS^{+}}$ according to Eqn.~(\ref{eqnsupp:self-attention-final}) and obtain adapted embeddings $\{\tilde{\V{u}}_{q,+}\}\cup\{\tilde{\V{u}}_{k,q}|v_k\in\CV_{\CS^{+}}\}$.
    \STATE Adapt $v_q$ together with $\CV_{\CS^{-}}$ according to Eqn.~(\ref{eqnsupp:self-attention-final}) and obtain adapted embeddings $\{\tilde{\V{u}}_{q,-}\}\cup\{\tilde{\V{u}}_{k,q}|v_k\in\CV_{\CS^{-}}\}$.
    \STATE Calculate the positive and negative prototypes $\tilde{\V{c}}_{m,q}, m\in\{+,-\}$ for classification according to Eqn.~(\ref{eqn:protos}).
    \STATE Calculate the predicted probability with $\tilde{\V{c}}_{m,q}$ and $\tilde{\V{u}}_{q,m}$ according to Eqn.~(\ref{eqn:prob}).
\ENDFOR

\end{algorithmic}
\end{algorithm}

\subsection{Baselines}

The following baselines are considered:

\noindent
\textbf{Label Propagation (LP)}~\cite{zhu2003semi}:
This method is a semi-supervised learning algorithm that estimates labels by propagating label information through a graph. It assigns a node the label which most of its neighborhoods have and propagates until no label is changing.

\noindent
\textbf{LINE}~\cite{tang2015line}:
This method first separately learns node embeddings by preserving 1- and 2-step neighborhood information between nodes and then concatenates them as the final node embeddings.

\noindent
\textbf{Node2Vec}~\cite{grover2016node2vec}:
This method converts graph structure to node sequences by mixing breadth- and depth-first random walk strategies and learns node embeddings with the skip-gram model~\cite{mikolov2013distributed}.

\noindent
\textbf{GCN}~\cite{kipf2016semi}:
This method is a semi-supervised method that uses a localized first-order approximation of spectral graph convolutions to exploit the graph structure.
Here we use the learned Node2Vec embeddings as the input feature matrix of GCN. %

\noindent
\textbf{Planetoid}~\cite{yang2016revisiting}:
This is a semi-supervised method that learns node embeddings by using them to
jointly predict node labels and node neighborhoods in the graph.

\noindent
\textbf{Meta-GNN}~\cite{zhou2019meta}:
This method directly applies MAML~\cite{finn2017model} to train GCN~\cite{kipf2016semi} in a meta-learning manner. Similarly, we use the learned Node2Vec embeddings as the input feature matrix of GCN.

\noindent
\textbf{Baseline Evaluation Procedure.}
We assess the performance of the baselines on the node classification tasks sampled from the test labels as follows:
(1) For LP, we propagate the labels of the support nodes over the entire graph and inspect the predicted labels of the query nodes for each test tasks;
(2) For each unsupervised network embedding method, we take the learned node embeddings as features to train a logistic regression classifier with L2 regularization for each test task. We use the support set to train the classifier and then predict the labels of the query nodes;
(3) For each semi-supervised network embedding method, we first use the training labels to train the model for multi-label node classification.
Then, for each test task, we fine-tune the model by substituting the final classification layer with a binary classification layer.
Analogous to (2), we use the support set to train the new layer and then predict the labels of the query nodes;
(4) For Meta-GNN, we first employ MAML~\cite{finn2017model} to learn a good initialization of GCN on the training tasks (binary node classification tasks).
Then, for each test task, we use the support set to update the GCN
from the learned initialization
and apply the adapted GCN to the query nodes.

\subsection{Parameter Settings}
For LP, we use an open-source implementation\ignorespaces
\footnote{\url{https://github.com/yamaguchiyuto/label\_propagation}}
and set the maximum iteration number to 30.
For fair comparisons, we set the dimension of node representations to 128
for LINE, Node2vec, and Planetoid.
For LINE,
we set the initial learning rate to 0.025 and the number of negative samples to 5.
For Node2vec,
we set the window size to 10, the length of each walk to 40, and the number of walks per node to 80.
The best in-out and return hyperparameters are tuned on the validation tasks with a grid search
over $p,q\in\{0.25,0.5,1,2,4\}$.
For Planetoid, we use the variant Planetoid-G since there are no input node features in our datasets.
We tune the respective batch sizes and learning rates used for optimizing the supervised and the structural objectives
based on the performance on the validation tasks.
For GCN, we use a two-layer GCN with the number of hidden units as 128 and ReLU nonlinearity,
and tune the dropout rate, learning rate, and weight decay based on the performance on the validation tasks.
and set other hyperparameters as the original paper.
For Meta-GNN\footnote{Since the authors do not provide the implementation that uses GCN as the learner, we implement it on the basis of the released code at \url{https://github.com/ChengtaiCao/Meta-GNN} to perform experiments.}, we also use a two-layer GCN with 128 hidden units and ReLU nonlinearity.
We set the number of inner updates to 2 due to the limitation of GPU memory
and tune the fast and meta learning rates based on the performance on the validation tasks.
For Planetoid, GCN, and Meta-GNN, we apply the best performing models on the validation tasks to the test tasks.

For our proposed \method{}, there are three parts of hyperparameters.
In the structural module,
we need to set the size $d$ of node representations and sample $N_1$ node pairs at each training step. We also sample $N_{\text{neg}}$ negative nodes per pair to speed up the calculation as in~\cite{tang2015line}.
In the meta-learning module, we sample $N_2$ training tasks at each training step.
The hyperparameters involved in the transformation function include
the number $H$ of parallel attention heads,
the size $d^\prime\!/H$ of the query, key, and value vectors,
the size $d_{\text{ff}}$ of the hidden layer in the two-layer feed-forward network,
the number $L$ of stacked computation blocks.
Besides, we apply dropout to the output of each of the self-attention modules and the feed-forward networks before it is added to the corresponding input and normalized,
and the dropout rate is denoted by $P_{\text{drop}}$.
Another hyperparameter is the weight decay coefficient $\lambda$.
In the optimization module,
we use the Adam optimizer~\cite{kingma2014adam} to optimize the structural and the meta-learning modules
with learning rates of $\alpha_1$ and $\alpha_2$, respectively.
In addition, we have the decay rate $\gamma$ and the decay period $N_{\text{decay}}$
to control the optimization of the structural and meta-learning modules.

For all four datasets, we set
$d=128$, $N_{\text{neg}}=5$, $P_{\text{drop}}=0.1$, and $\gamma=0.1$.
We tune other hyperparameters on the validation tasks over the search space shown in Table~\ref{tab:search}.
We utilize the Ray Tune library~\cite{liaw2018tune} with asynchronous HyperBand scheduler~\cite{li2018massively}
to accelerate the searching process.
Note that, for each dataset, we only search the best hyperparameters
with $K_{\ast,+}=10$ and $K_{\ast,-}=20$ for both training and test tasks,
and directly apply these hyperparameters to
other experimental scenarios.
The resulting hyperparameters are available in our attached code.

\begin{table}[t]
\centering
\caption{The hyperparameter search space.}
\label{tab:search}
\begin{tabular}{@{}cc|cc@{}}
\toprule
Hyperparameter & Values & Hyperparameter & Values\\ \midrule
$N_1$ & $\{512,1024,2048\}$ & $L$ & $\{1,2,3\}$ \\
$N_2$ & $\{32,64,128\}$ & $\lambda$ & $\{0.001,0.01,0.1\}$ \\
$H$ & $\{1,2,4\}$ & $\alpha_1$ & $\{0.0001,0.001\}$ \\
$d^\prime$ & $\{128,256\}$ & $\alpha_2$ & $\{0.0001,0.001\}$ \\
$d_{\text{ff}}$ & $\{256,512\}$ & $N_{\text{decay}}$ & $\{500,1000,1500,2000\}$ \\ \bottomrule
\end{tabular}
\end{table}

\section{Additional Experiments}

\subsection{Full Results of Overall Comparisons}

The full results of overall comparisons in our original paper
are presented in Table~\ref{tab:expt_supple} in the form of $\mathrm{mean \pm std}$.
Overall, we observe that our proposed \method{} achieves comparable or even lower standard deviation,
which demonstrates the statistical significance of the superiority of \method{}.

\begin{table*}[t]
\centering
\caption{Results with standard deviation on few-shot node classification tasks with novel labels. OOM means out of memory (16 GB GPU memory).}
\label{tab:expt_supple}
\subcaptionbox{$K_{\ast,+}=10$ and $K_{\ast,-}=20$.}{
\resizebox{0.98\linewidth}{!}{%
\begin{tabular}{@{}lrrrrrr@{}}
\toprule
\multirow{2}{*}{Method} & \multicolumn{3}{c}{BlogCatalog} & \multicolumn{3}{c}{Flickr}   \\ \cmidrule(lr){2-4} \cmidrule(lr){5-7}
 & \multicolumn{1}{c}{AUC} & \multicolumn{1}{c}{F$_1$} & \multicolumn{1}{c}{Recall} & \multicolumn{1}{c}{AUC} & \multicolumn{1}{c}{F$_1$} & \multicolumn{1}{c}{Recall}  \\ \midrule
LP & 0.6422$^{\pm 0.0289}$ & 0.1798$^{\pm0.0198}$ & 0.2630$^{\pm0.0309}$ & 0.8196$^{\pm0.0175}$ & 0.4321$^{\pm0.0392}$ & 0.4989$^{\pm0.0492}$  \\
LINE & 0.6690$^{\pm0.0323}$ & 0.2334$^{\pm0.0499}$ & 0.1595$^{\pm0.0403}$ & 0.8593$^{\pm0.0145}$ & 0.6194$^{\pm0.0334}$ & 0.5418$^{\pm0.0382}$ \\
Node2vec & 0.6697$^{\pm0.0325}$ & 0.3750$^{\pm0.0478}$ & 0.2940$^{\pm0.0432}$ & 0.8504$^{\pm0.0151}$ & 0.6664$^{\pm0.0284}$ & 0.6147$^{\pm0.0332}$  \\

Planetoid & 0.6850$^{\pm0.0320}$ & 0.4657$^{\pm0.0437}$ & 0.4301$^{\pm0.0451}$ & \textbf{0.8601}$^{\pm0.0360}$ & 0.6638$^{\pm0.0796}$ &0.6331$^{\pm0.0821}$   \\

GCN & 0.6643$^{\pm0.0288}$ & 0.3892$^{\pm0.0423}$ & 0.3379$^{\pm0.0401}$ & OOM & OOM & OOM  \\
Meta-GNN & 0.6533$^{\pm0.0362}$ & 0.3567$^{\pm0.0364}$ & 0.2962$^{\pm0.0398}$ & OOM & OOM & OOM \\ \midrule
\method{} & \textbf{0.6986}$^{\pm0.0305}$ & \textbf{0.5380}$^{\pm0.0342}$ & \textbf{0.6203}$^{\pm0.0375}$ & 0.8462$^{\pm0.0164}$ & \textbf{0.7118}$^{\pm0.0223}$ & \textbf{0.7700}$^{\pm0.0227}$   \\
\%Improv. & 1.99 & 15.53 & 44.22 & -1.62 & 6.81 & 21.62  \\ \bottomrule
\end{tabular}%
}
}\\

\subcaptionbox*{\vspace{-10pt}}{
\resizebox{0.98\linewidth}{!}{%
\begin{tabular}{@{}lrrrrrr@{}}
\toprule
\multirow{2}{*}{Method} & \multicolumn{3}{c}{PPI} & \multicolumn{3}{c}{Mashup}   \\ \cmidrule(lr){2-4} \cmidrule(lr){5-7}
 & \multicolumn{1}{c}{AUC} & \multicolumn{1}{c}{F$_1$} & \multicolumn{1}{c}{Recall} & \multicolumn{1}{c}{AUC} & \multicolumn{1}{c}{F$_1$} & \multicolumn{1}{c}{Recall}  \\ \midrule
LP & 0.6285$^{\pm0.0221}$ & 0.2147$^{\pm0.0384}$ & 0.2769$^{\pm0.0630}$ & 0.6488$^{\pm0.0258}$ & 0.3103$^{\pm0.0414}$ & 0.4535$^{\pm0.0991
}$   \\
LINE & 0.6372$^{\pm0.0270}$ & 0.2147$^{\pm0.0373}$ & 0.1456$^{\pm0.0280}$ & 0.6926$^{\pm0.0354}$ & 0.2970$^{\pm0.0602}$ & 0.2142$^{\pm0.0537}$ \\
Node2vec & 0.6273$^{\pm0.0258}$ & 0.3545$^{\pm0.0350}$ & 0.2860$^{\pm0.0326}$ & 0.6575$^{\pm0.0303}$ & 0.3835$^{\pm0.0413}$ & 0.3147$^{\pm0.0396}$ \\

Planetoid & 0.6791$^{\pm0.0251}$ & 0.4672$^{\pm0.0314}$ & 0.4411$^{\pm0.0328}$ & 0.7056$^{\pm0.0223}$ & 0.4825$^{\pm0.0287}$ & 0.4218$^{\pm0.0334}$  \\

GCN &  0.6596$^{\pm0.0223}$ & 0.4176$^{\pm0.0335}$ & 0.3729$^{\pm0.0327}$ & 0.6910$^{\pm0.0248}$ & 0.4065$^{\pm0.0417}$ & 0.3607$^{\pm0.0396}$  \\
Meta-GNN & 0.6537$^{\pm0.0307}$ & 0.3964$^{\pm0.0343}$ & 0.3373$^{\pm0.0405}$ & 0.7093$^{\pm0.0317}$ & 0.4689$^{\pm0.0389}$ & 0.4202$^{\pm0.0384}$  \\
\midrule

\method{} & \textbf{0.6865}$^{\pm0.0205}$ & \textbf{0.5188}$^{\pm0.0209}$ & \textbf{0.5621}$^{\pm0.0311}$ & \textbf{0.7645}$^{\pm0.0251}$ & \textbf{0.5764}$^{\pm0.0291}$ & \textbf{0.5566}$^{\pm0.0337}$  \\
\%Improv. & 1.09 & 11.04 & 27.43 & 7.78 & 19.46 & 22.73 \\ \bottomrule
\end{tabular}%
}
}\\

\vspace{10pt}

\subcaptionbox{$K_{\ast,+}=10$ and $K_{\ast,-}=40$.}{
\resizebox{0.98\linewidth}{!}{%
\begin{tabular}{@{}lrrrrrr@{}}
\toprule
\multirow{2}{*}{Method} & \multicolumn{3}{c}{BlogCatalog} & \multicolumn{3}{c}{Flickr}   \\ \cmidrule(lr){2-4} \cmidrule(lr){5-7}
 & \multicolumn{1}{c}{AUC} & \multicolumn{1}{c}{F$_1$} & \multicolumn{1}{c}{Recall} & \multicolumn{1}{c}{AUC} & \multicolumn{1}{c}{F$_1$} & \multicolumn{1}{c}{Recall}  \\ \midrule
LP &0.6421$^{\pm0.0288}$ & 0.0554 $^{\pm0.0118}$& 0.0727$^{\pm0.0158}$ & 0.8253$^{\pm0.0156}$ & 0.3055$^{\pm0.0413}$ & 0.3040$^{\pm0.0485}$  \\
LINE & 0.6793$^{\pm0.0320}$ & 0.0529$^{\pm0.0316}$ & 0.0328$^{\pm0.0216}$ & 0.8644$^{\pm0.0139}$ & 0.4154$^{\pm0.0471}$ & 0.3485$^{\pm0.0471}$ \\
Node2vec & 0.6792$^{\pm0.0314}$ & 0.1982$^{\pm0.0516}$ & 0.1340$^{\pm0.0398}$ & 0.8558$^{\pm0.0150}$ & 0.5295$^{\pm0.0381}$ & 0.4602$^{\pm0.0420}$  \\

Planetoid & 0.6981$^{\pm0.0315}$ & 0.2980$^{\pm0.0550}$ & 0.2319$^{\pm0.0507}$ & \textbf{0.8728}$^{\pm0.0382}$ & 0.5040$^{\pm0.0790}$ & 0.4461$^{\pm0.0741}$   \\

GCN & 0.6794$^{\pm0.0302}$ & 0.2104$^{\pm0.0347}$ & 0.1583$^{\pm0.0268}$ & OOM & OOM & OOM \\
Meta-GNN & 0.6724$^{\pm0.0396}$ & 0.2152$^{\pm0.0578}$ & 0.1618$^{\pm0.0546}$ & OOM & OOM & OOM \\ \midrule
\method{} & \textbf{0.7139}$^{\pm0.0309}$ & \textbf{0.4398}$^{\pm0.0401}$ & \textbf{0.5819}$^{\pm0.0451}$ & 0.8505$^{\pm0.0154}$ & \textbf{0.6220}$^{\pm0.0245}$ & \textbf{0.7460}$^{\pm0.0523}$  \\
\%Improv. & 2.26 & 47.58 & 150.93 & -2.55 & 17.47 & 62.10 \\ \bottomrule
\end{tabular}%
}
}\\

\subcaptionbox*{\vspace{-10pt}}{
\resizebox{0.98\linewidth}{!}{%
\begin{tabular}{@{}lrrrrrr@{}}
\toprule
\multirow{2}{*}{Method} & \multicolumn{3}{c}{PPI} & \multicolumn{3}{c}{Mashup}   \\ \cmidrule(lr){2-4} \cmidrule(lr){5-7}
 & \multicolumn{1}{c}{AUC} & \multicolumn{1}{c}{F$_1$} & \multicolumn{1}{c}{Recall} & \multicolumn{1}{c}{AUC} & \multicolumn{1}{c}{F$_1$} & \multicolumn{1}{c}{Recall}  \\ \midrule
LP & 0.6298$^{\pm0.0228}$ & 0.0773$^{\pm0.0231}$ & 0.0748$^{\pm0.0277}$ & 0.6534$^{\pm0.0259}$ & 0.1156$^{\pm0.0276}$ & 0.1284$^{\pm0.0509}$   \\
LINE & 0.6423$^{\pm0.0268}$ & 0.0496$^{\pm0.0193}$ & 0.0300$^{\pm0.0122}$ & 0.7009$^{\pm0.0345}$ & 0.0956$^{\pm0.0489}$ & 0.0617$^{\pm0.0348}$\\
Node2vec & 0.6309$^{\pm0.0264}$ & 0.1894$^{\pm0.0373}$ & 0.1306$^{\pm0.0286}$ & 0.6643$^{\pm0.0311}$ & 0.2070$^{\pm0.0417}$ & 0.1447$^{\pm0.0333}$ \\

Planetoid & 0.6879$^{\pm0.0250}$ & 0.3100$^{\pm0.0368}$ & 0.2523$^{\pm0.0323}$ & 0.7095$^{\pm0.0223}$ & 0.3279$^{\pm0.0298}$ & 0.2551$^{\pm0.0278}$ \\

GCN &  0.6608$^{\pm0.0225}$ & 0.2531$^{\pm0.0353}$ & 0.1974$^{\pm0.0268}$ & 0.7007$^{\pm0.0245}$ & 0.2558$^{\pm0.0237}$ & 0.2098$^{\pm0.0169}$ \\
Meta-GNN & 0.6617$^{\pm0.0309}$ & 0.2575$^{\pm0.0332}$ & 0.2088$^{\pm0.0396}$ & 0.7140$^{\pm0.0339}$ & 0.3412$^{\pm0.0554}$ & 0.2864$^{\pm0.0635}$  \\
\midrule

\method{} & \textbf{0.7039}$^{\pm0.0218}$ & \textbf{0.4298}$^{\pm0.0242}$ & \textbf{0.5327}$^{\pm0.0420}$ & \textbf{0.7684}$^{\pm0.0244}$ & \textbf{0.4814}$^{\pm0.0318}$ & \textbf{0.4816}$^{\pm0.0393}$  \\
\%Improv. & 2.33 & 38.65 & 111.14 & 7.62 & 41.09 & 68.16 \\ \bottomrule
\end{tabular}%
}
}\\

\end{table*}

\subsection{The Performance w.r.t. the Numbers of Positive and Negative Nodes}

\begin{figure}[h]
\centering
\begin{subfigure}[b]{0.37\linewidth}
\includegraphics[width=\linewidth]{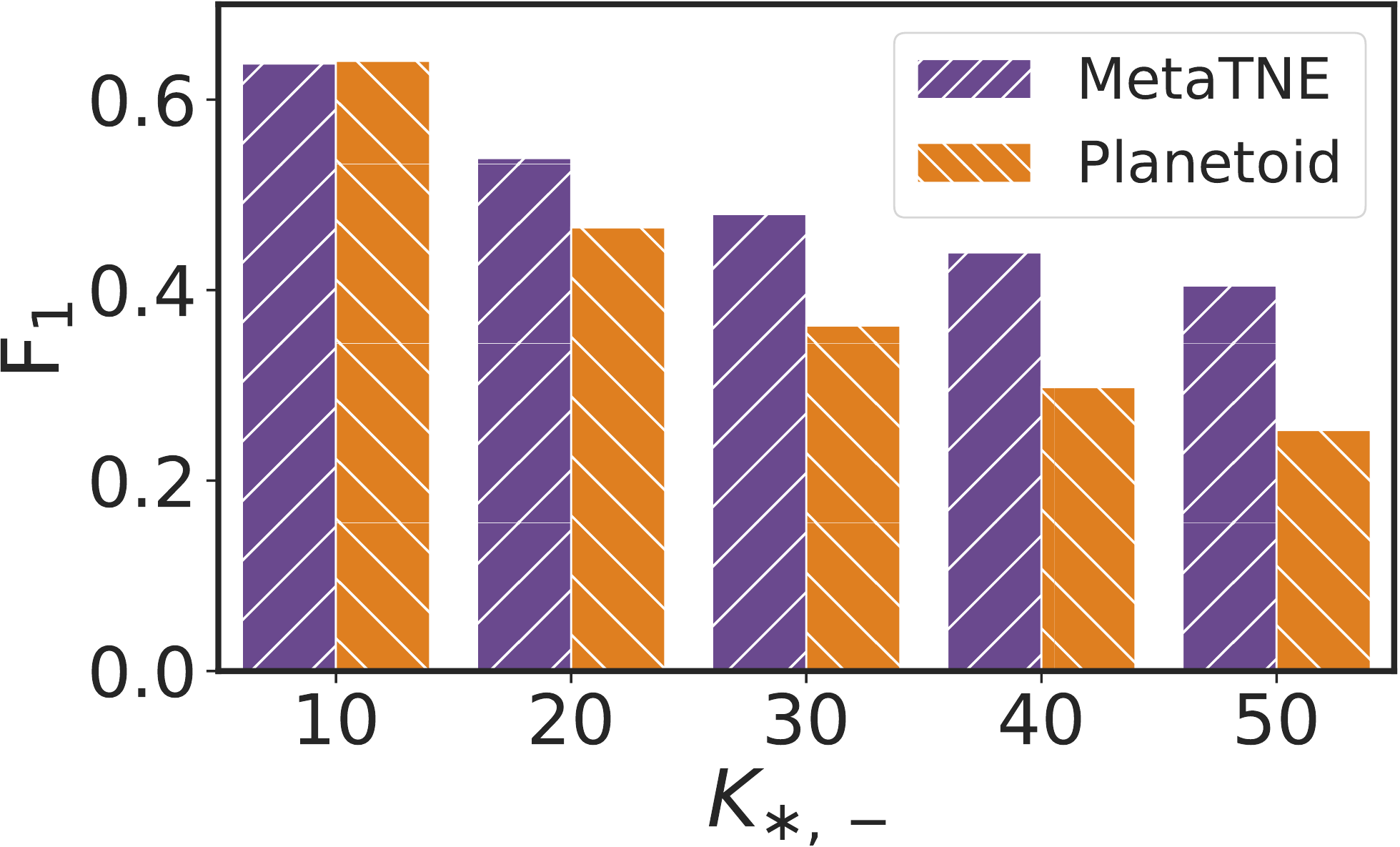}
\caption{$K_{\ast,+}=10$.}
\end{subfigure}
\hspace{15pt}
\begin{subfigure}[b]{0.37\linewidth}
\includegraphics[width=\linewidth]{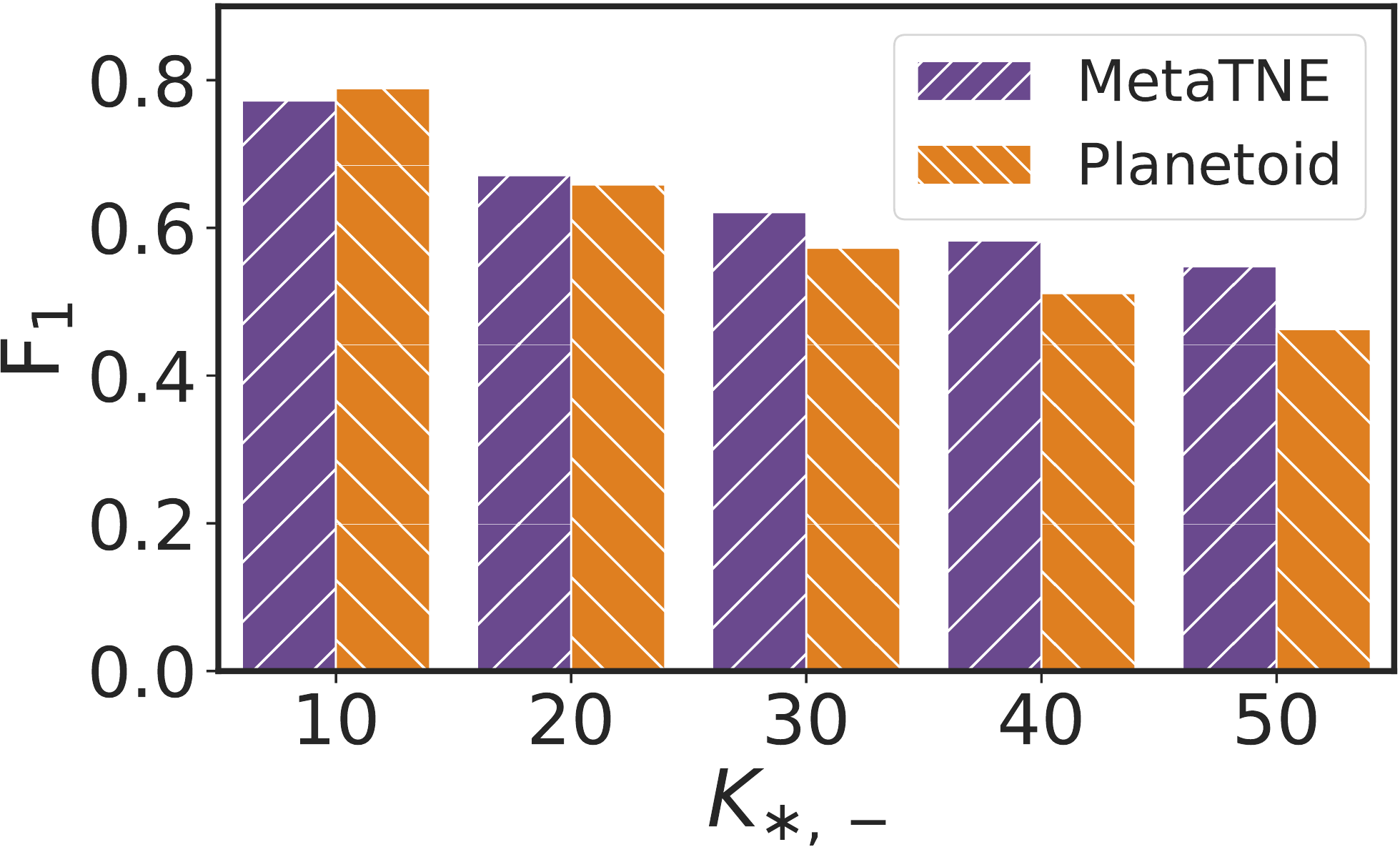}
\caption{$K_{\ast,+}=20.$}
\end{subfigure}
\caption{The performance w.r.t. the numbers of positive and negative nodes on BlogCatalog dataset.}
\label{fig:varying_neg}
\end{figure}

To further investigate the performance
under different combinations of $K_{\ast,+}$ and $K_{\ast,-}$,
we conduct experiments with $K_{\ast,+}$ fixed at either 10 or 20 while varying $K_{\ast,-}$ from 10 to 50 for both training and test tasks.
Figure~\ref{fig:varying_neg} gives the performance comparisons of \method{} and the best performing baseline
(i.e., Planetoid) in terms of F$_1$ on BlogCatalog dataset.
We observe that Planetoid and \method{} achieve comparable performance
when $K_{\ast,+}$ is the same as or larger than $K_{\ast,-}$,
while the performance gap between \method{} and Planetoid gradually increases
as the ratio of $K_{\ast,+}$ to $K_{\ast,-}$ decreases,
which demonstrates the practicability of our method since the positive nodes are relatively scarce compared with the negative ones in many realistic applications.

\subsection{The Performance w.r.t. the Number of Query Nodes}\label{sec:varying_query}

\begin{figure}[t]
\centering
\begin{subfigure}[b]{0.4\linewidth}
\includegraphics[width=\linewidth]{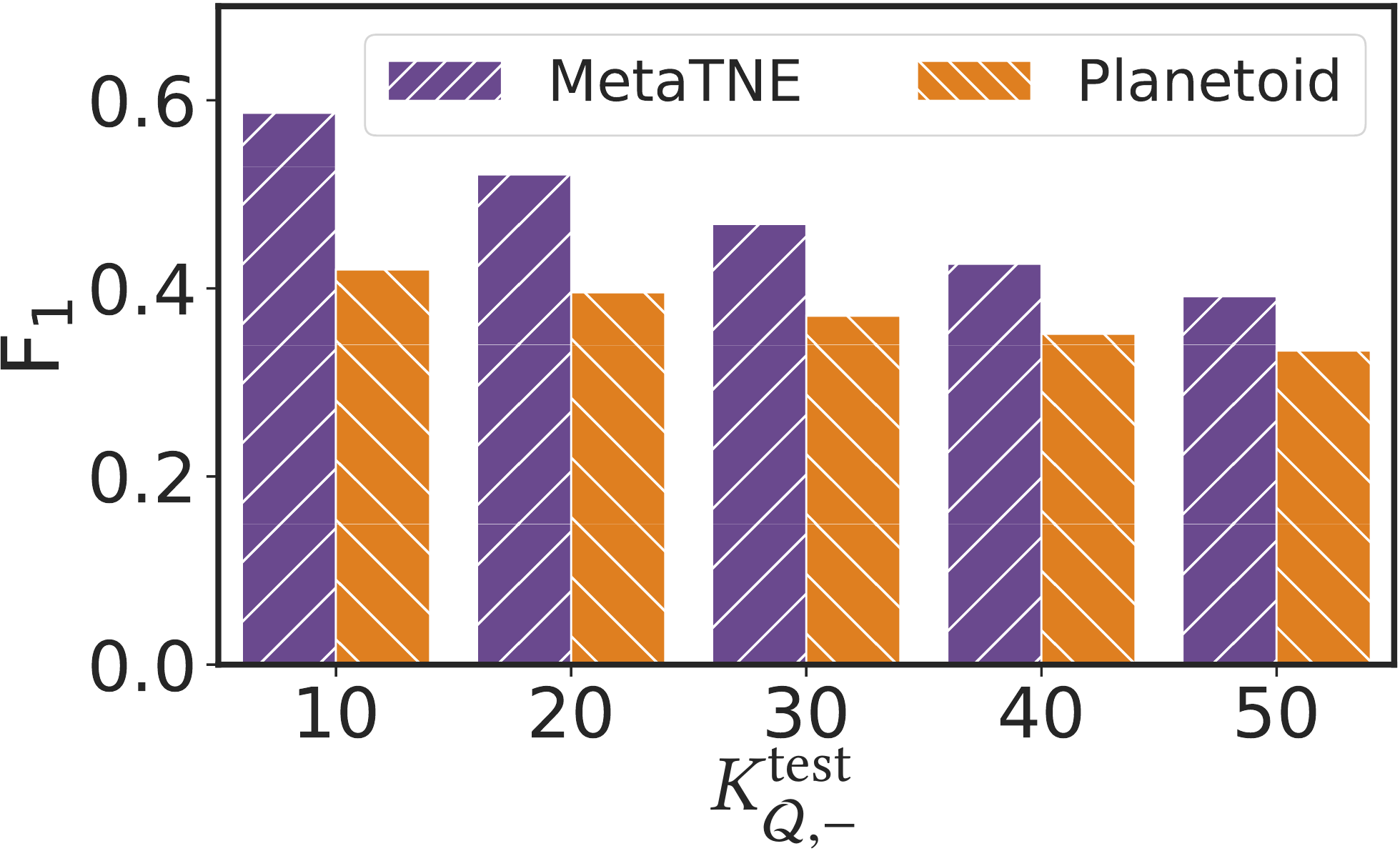}
\caption{$K_{\CQ,+}^{\mathrm{test}}=10$.}\label{fig:varying_query_10}
\end{subfigure}
\hspace{12pt}
\begin{subfigure}[b]{0.4\linewidth}
\includegraphics[width=\linewidth]{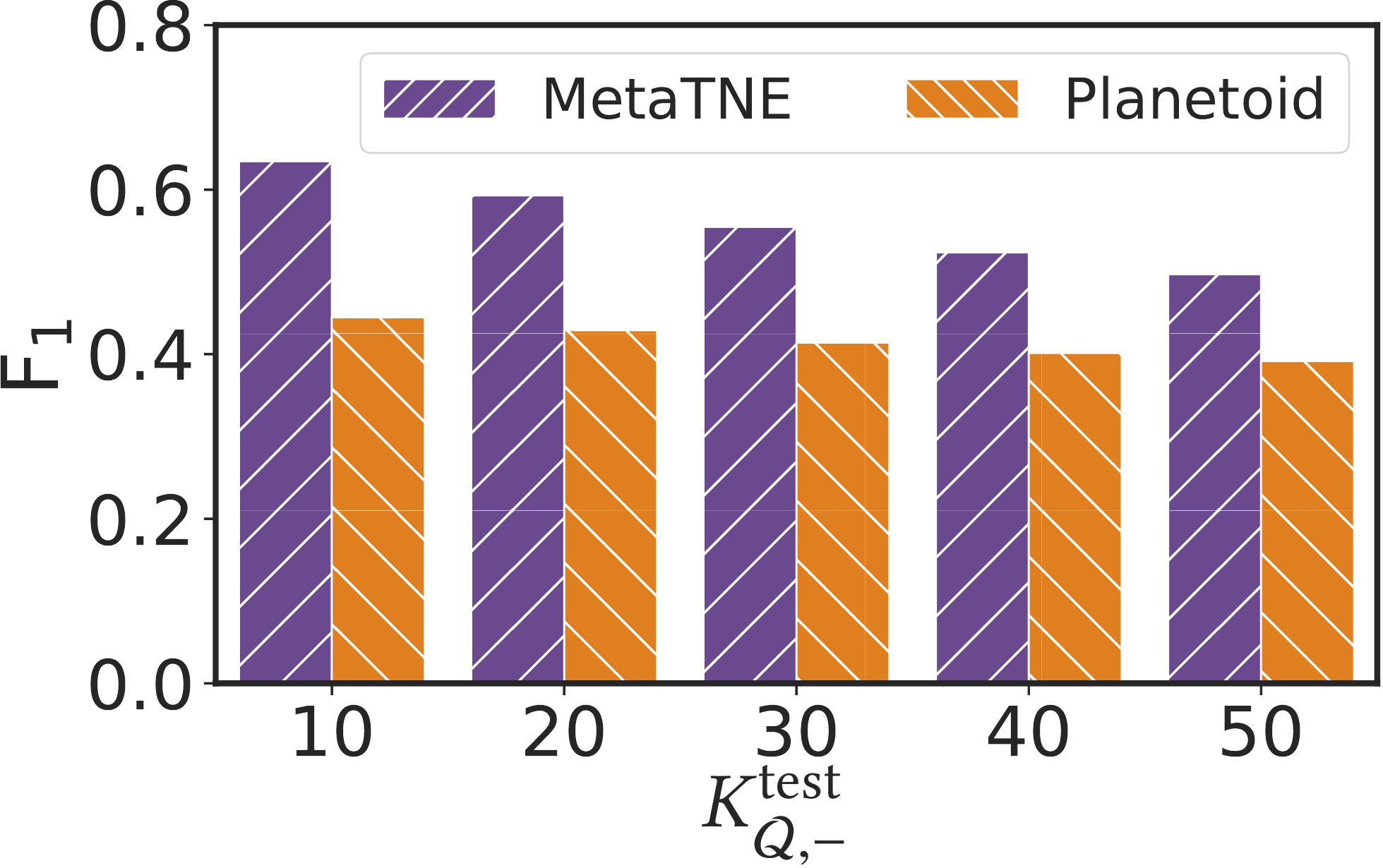}
\caption{$K_{\CQ,+}^{\mathrm{test}}=20.$}
\end{subfigure}
\caption{The performance w.r.t. the number of query nodes on PPI dataset.}
\label{fig:varying_query}
\end{figure}

In the above experiments, we presume that, for each few-shot node classification task,
the support and the query sets have the same numbers of positive and negative nodes
following the standard protocol of meta-learning (called the \emph{standard-setting}).
However, in practice, the query set could have different numbers of positive and negative nodes
as well as a different ratio of the number of positive nodes to the number of negative nodes compared to the support set.
Thus, we further examine how the number of query nodes influences the performance.
Towards this end,
we sample additional test tasks by varying the numbers of positive and negative nodes in the query set
(i.e., $K_{\CQ,+}^{\text{test}}$ and $K_{\CQ,-}^{\text{test}}$),
with the numbers of positive and negative nodes in the support set fixed at 10 and 30, respectively (i.e., $K_{\CS,+}^{\text{test}}=10$ and $K_{\CS,-}^{\text{test}}=30$),
and then compare the performance on these tasks.
This setting is called the \emph{generalized-setting}.
Note that here we only alter the sampling of test tasks as described above
and the training tasks are always sampled under the condition that both the support and query sets
contain 10 positive and 30 negative nodes (i.e., $K_{\ast,+}^{\text{train}}=10$ and $K_{\ast,-}^{\text{train}}=30$).
Figure~\ref{fig:varying_query} shows the experimental results on PPI dataset.

We observe that \method{} consistently yields better performance than Planetoid
under different combinations of $K_{\CQ,+}^{\text{test}}$ and $K_{\CQ,-}^{\text{test}}$.
In particular,
jointly analyzing Table~\ref{tab:expt_supple} and Fig.~\ref{fig:varying_query_10},
\method{} achieves almost the same performance in both the standard- and generalized-settings
when the query set contains 10 positive nodes as well as 20 or 40 negative nodes,
which indicates that to some extent
\method{} is not sensitive to the choice of $K_{\ast,+}$ and $K_{\ast,-}$ for sampling training tasks
as well as $K_{\CS,+}^{\text{test}}$ and $K_{\CS,+}^{\text{test}}$
and demonstrates the robustness of \method{}.
On the other hand,
it essentially becomes easier to classify the query nodes
as the ratio of $K_{\CQ,+}^{\text{test}}$ to $K_{\CQ,-}^{\text{test}}$ increases,
whereas the performance of Planetoid does not change markedly
as $K_{\CQ,-}^{\text{test}}$ decreases
in Fig.~\ref{fig:varying_query},
which evidences that Planetoid tends to overfit
the training tasks (e.g., the ratio of the number of positive nodes to the number of negative nodes).

\subsection{The Performance with Fewer Positive Nodes}

We further examine the performance of different methods by using fewer positive nodes
and
conduct experiments with $K_{\ast,+}$ set to $5$ and $K_{\ast,-}$ set to $10$ or $20$.
Table~\ref{table:fewer_positive} reports the experimental results on BlogCatalog dataset.
From Table~\ref{table:fewer_positive},
we observe similar results to Table~\ref{tab:expt_supple} and \method{} still significantly outperforms
all other methods in the case that there are fewer positive nodes.

\begin{table}[!h]
\centering
\caption{Results of fewer positive nodes on BlogCatalog dataset.}
\label{table:fewer_positive}
\resizebox{0.98\linewidth}{!}{%
\begin{tabular}{@{}lrrrrrr@{}}
\toprule
\multirow{2}{*}{Method} & \multicolumn{3}{c}{$K_{\ast,+}=5,K_{\ast,-}=10$} & \multicolumn{3}{c}{$K_{\ast,+}=5,K_{\ast,-}=20$} \\ \cmidrule(lr){2-4} \cmidrule(lr){5-7}
 & \multicolumn{1}{c}{AUC} & \multicolumn{1}{c}{F$_1$} & \multicolumn{1}{c}{Recall} & \multicolumn{1}{c}{AUC} & \multicolumn{1}{c}{F$_1$} & \multicolumn{1}{c}{Recall} \\ \midrule
LP & 0.6231$^{\pm 0.0284}$ & 0.1753$^{\pm 0.0168}$ & 0.2831$^{\pm 0.0279}$ & 0.6226$^{\pm 0.0288}$ & 0.0567$^{\pm 0.0101}$ & 0.0930$^{\pm 0.0159}$ \\
LINE & 0.6355$^{\pm 0.0295}$ & 0.1296$^{\pm 0.0379}$ & 0.0884$^{\pm 0.0291}$ & 0.6432$^{\pm 0.0300}$ & 0.0116$^{\pm 0.0141}$ & 0.0076$^{\pm 0.0098}$ \\
Node2vec & 0.6384$^{\pm 0.0299}$ & 0.2912$^{\pm 0.0440}$ & 0.2267$^{\pm 0.0387}$ & 0.6451$^{\pm 0.0305}$ & 0.1017$^{\pm 0.0372}$ & 0.0689$^{\pm 0.0273}$ \\
Planetoid & 0.6473$^{\pm 0.0303}$ & 0.4221$^{\pm 0.0408}$ & 0.4052$^{\pm 0.0437}$ & 0.6583$^{\pm 0.0318}$ & 0.2305$^{\pm 0.0509}$ & 0.1853$^{\pm 0.0470}$ \\
GCN & 0.6379$^{\pm 0.0308}$ & 0.3376$^{\pm 0.0473}$ & 0.3015$^{\pm 0.0455}$ & 0.6524$^{\pm 0.0312}$ & 0.1590$^{\pm 0.0492}$ & 0.1239$^{\pm0.0408}$ \\
Meta-GNN & 0.6392$^{\pm 0.0362}$ & 0.3523$^{\pm 0.0375}$ &0.3152$^{\pm0.0468}$  & 0.6552$^{\pm 0.0399}$ &  0.1719$^{\pm 0.0612}$&  0.1485$^{\pm 0.0598}$\\ \midrule
MetaTNE & \textbf{0.6546}$^{\pm   0.0286}$ & \textbf{0.4523}$^{\pm 0.0371}$ & \textbf{0.4842}$^{\pm 0.0469}$ & \textbf{0.6756}$^{\pm   0.0295}$ & \textbf{0.3730}$^{\pm 0.0387}$ & \textbf{0.4539}$^{\pm 0.0505}$ \\
\%Improv. &1.13  & 7.15 &  19.50&2.63  & 61.82 &144.95  \\ \bottomrule
\end{tabular}%
}
\end{table}

\subsection{Visualization}\label{sec:visual}

\begin{figure}[ht]
\centering
\begin{subfigure}[b]{\linewidth}
\centering
\includegraphics[width=0.35\linewidth]{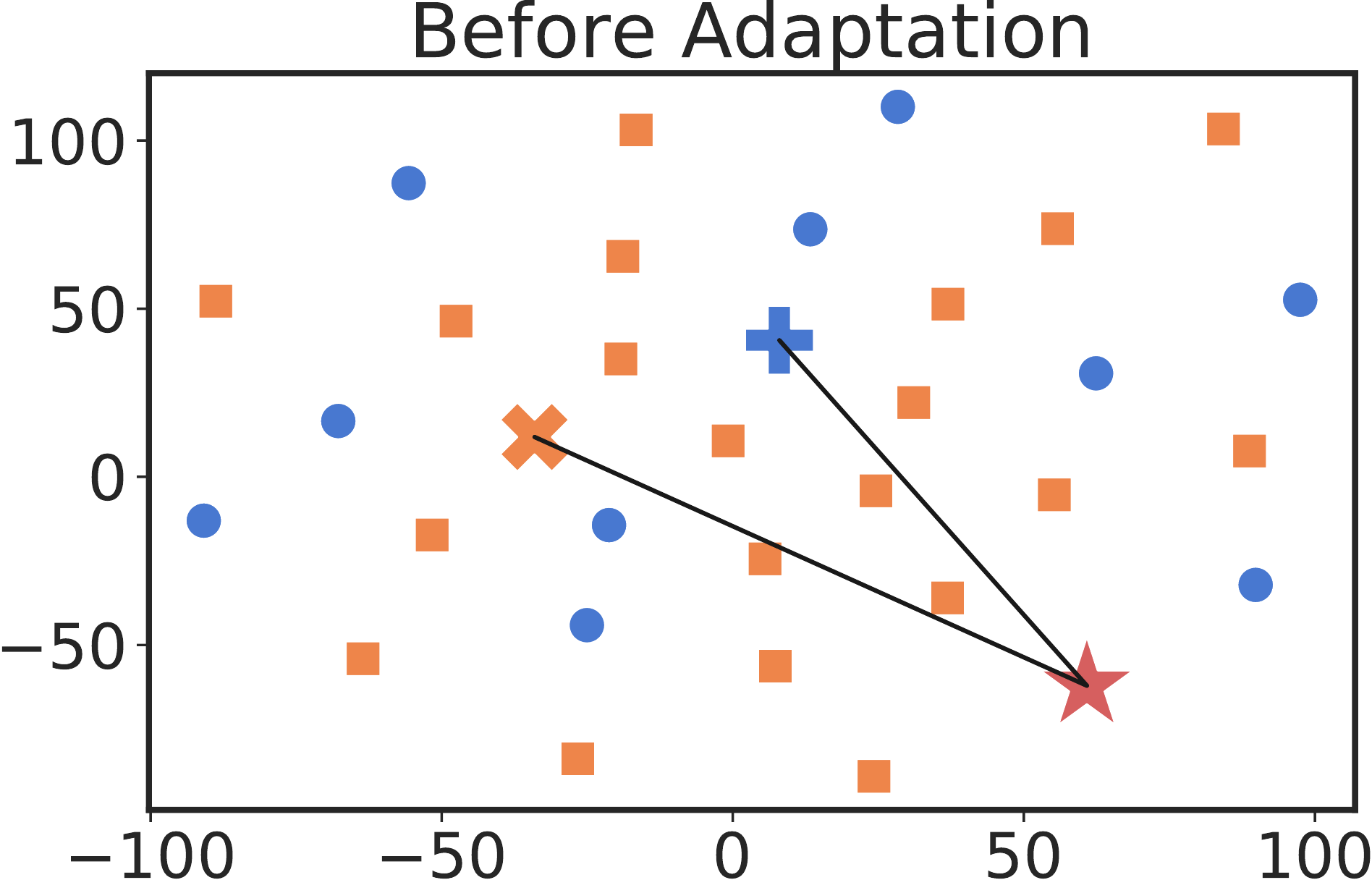}
\hspace{15pt}
\includegraphics[width=0.35\linewidth]{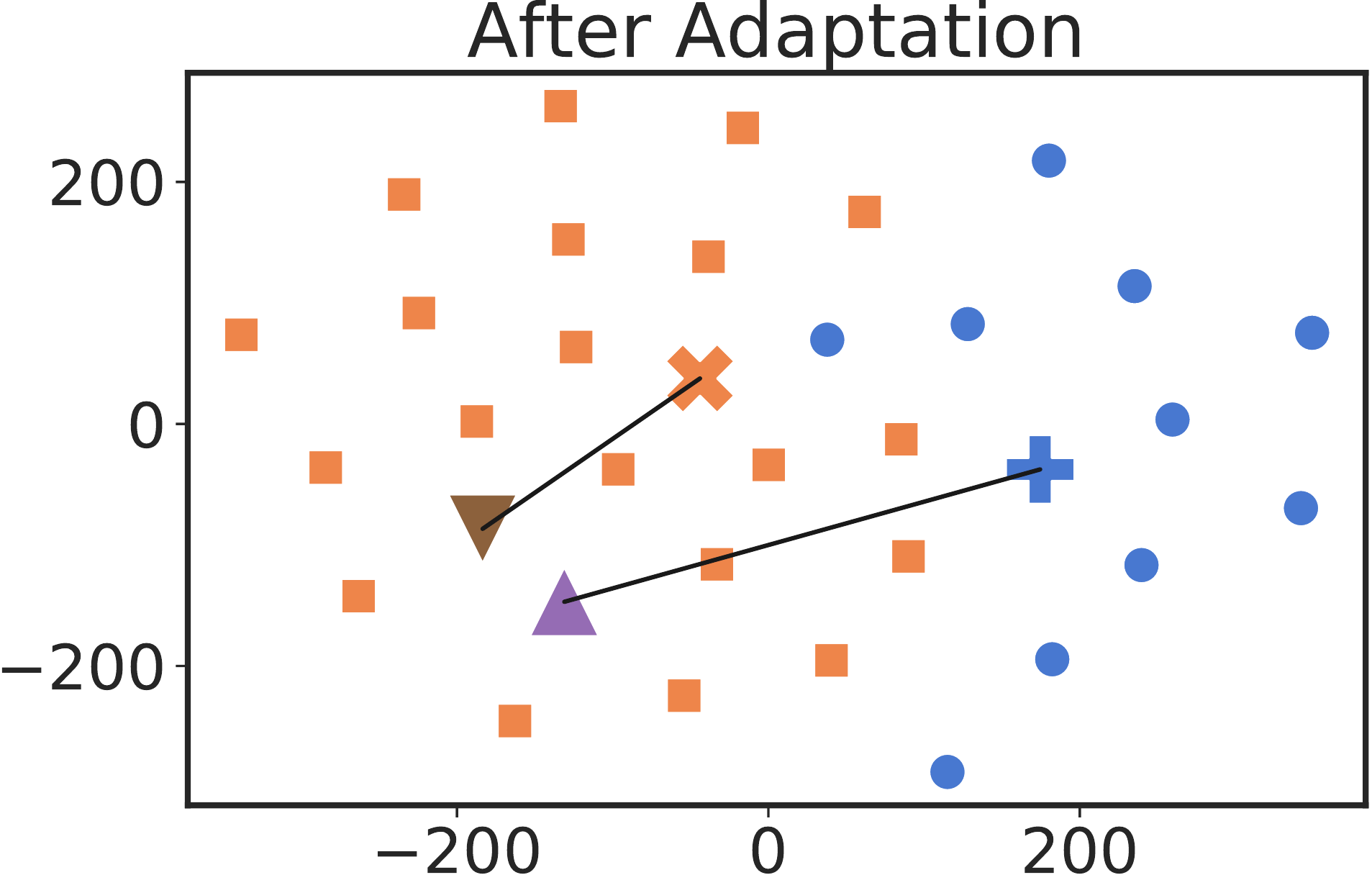}
\caption{The ground-truth of the query node is negative.}\label{fig:visual_1}
\vspace{5pt}
\end{subfigure}
\begin{subfigure}[b]{\linewidth}
\centering
\includegraphics[width=0.35\linewidth]{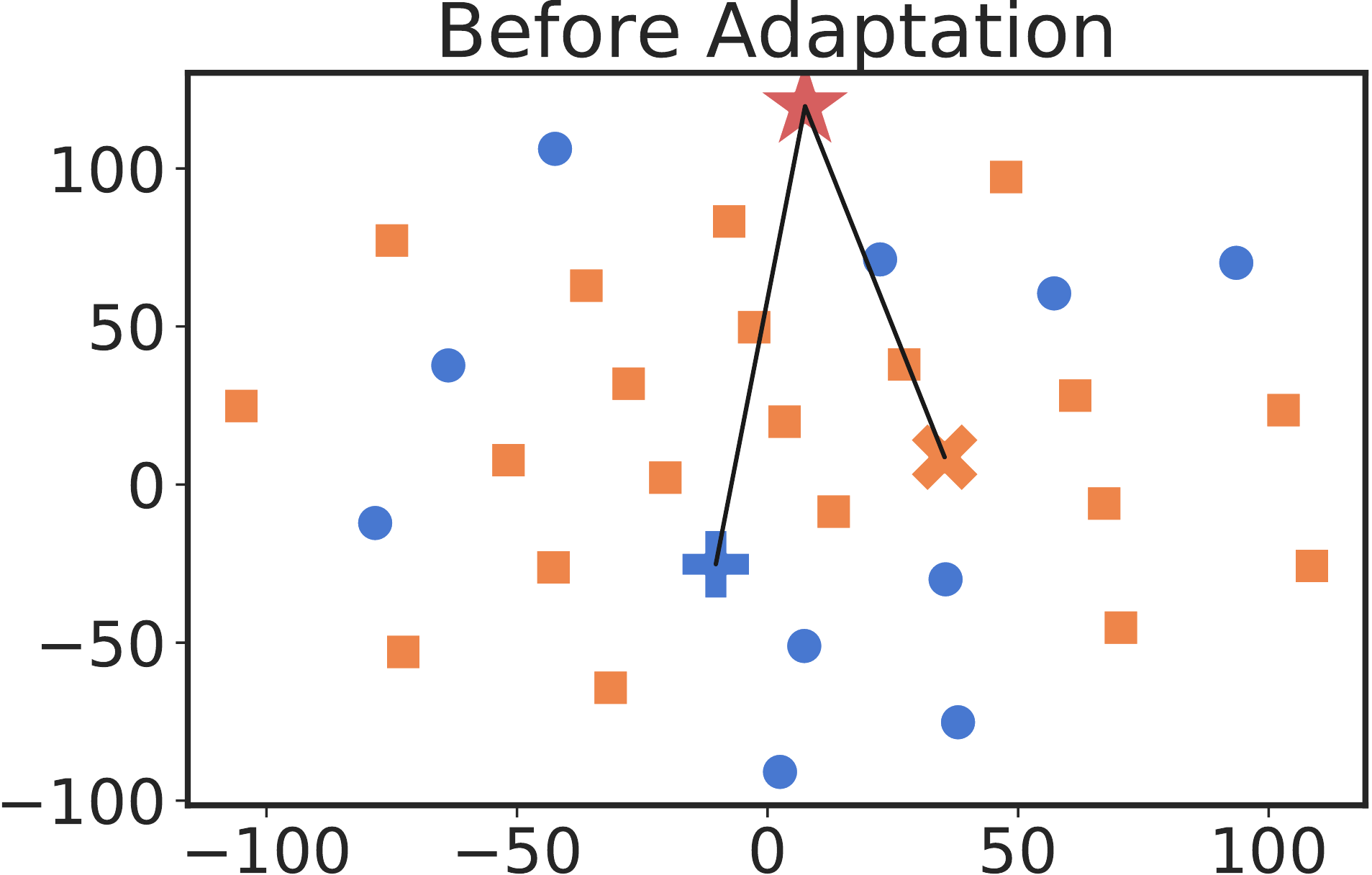}
\hspace{15pt}
\includegraphics[width=0.35\linewidth]{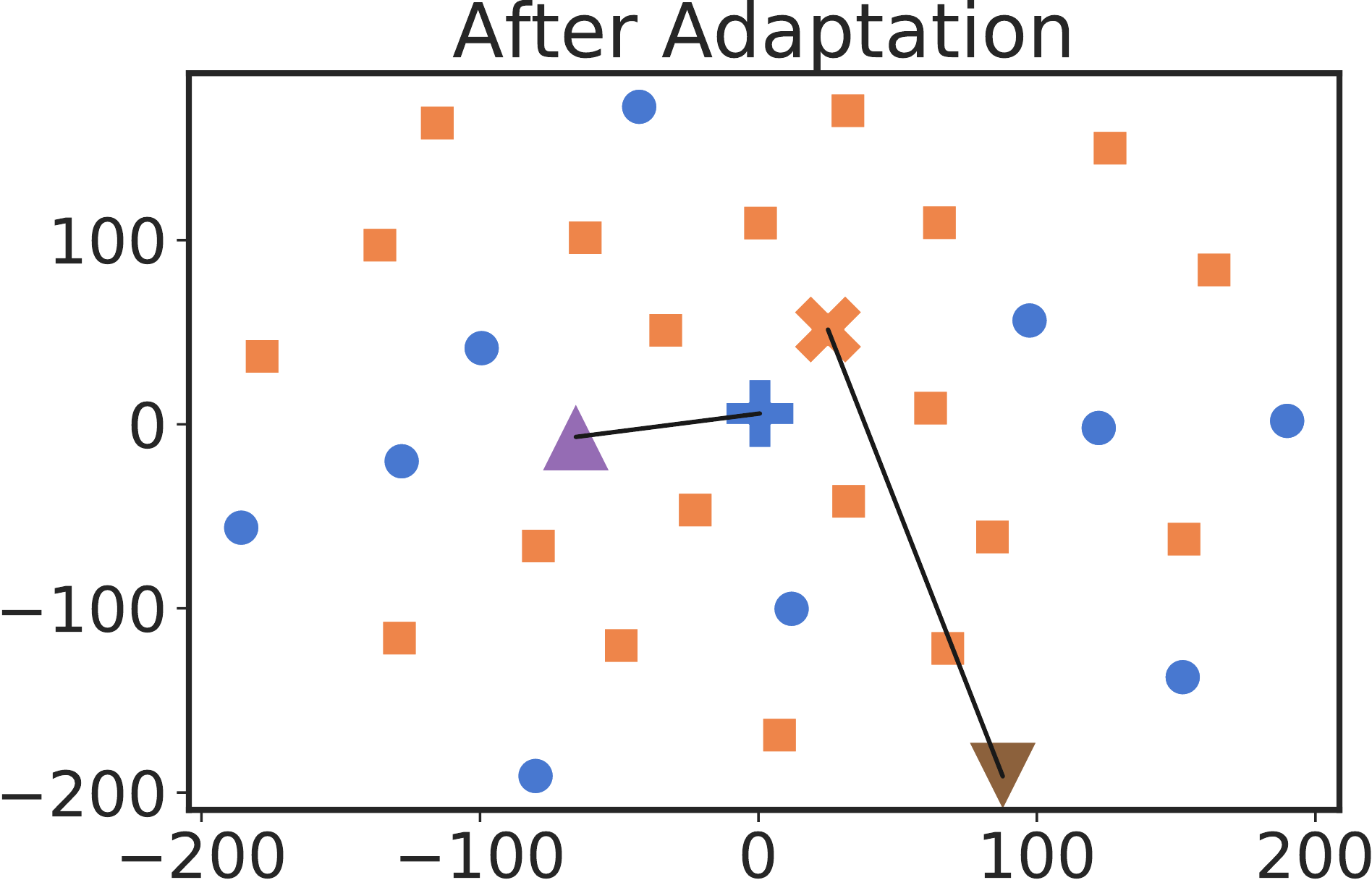}
\caption{The ground-truth of the query node is positive.}\label{fig:visual_2}
\vspace{3pt}
\end{subfigure}
\begin{subfigure}[b]{\linewidth}
\centering
\includegraphics[width=0.7\linewidth]{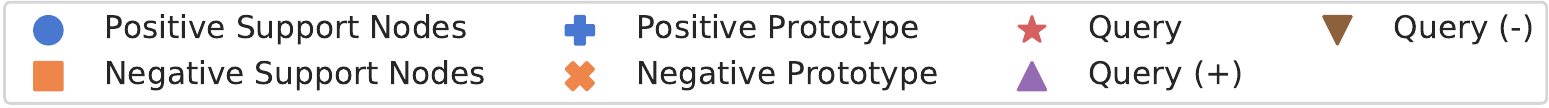}
\end{subfigure}
\caption{t-SNE visualization of embedding adaptation.}
\label{fig:visual}
\end{figure}

To better demonstrate the effectiveness of the transformation function,
we select two typical query nodes from the test tasks on Flickr dataset
and visualize the relevant node embeddings before and after adaptation
with t-SNE~\cite{maaten2008visualizing} in Fig.~\ref{fig:visual}.
Note that ``Query (+)'' and ``Query (-)'', respectively, indicate the adapted embeddings of the query node in relation to the positive and
negative support nodes in Eqn.~(\ref{eqn:self-attention-query}).
From Fig.~\ref{fig:visual_1} where the label of the query node is negative,
we see that, before adaptation, the embedding of the query node is closer to the positive prototype than the negative prototype and thus misclassification occurs.
After adaptation, the distance between ``Query (-)'' and the negative prototype is smaller than that between ``Query (+)'' and the positive prototype
and hence the query node is classified correctly.
The similar behavior is observed in Fig.~\ref{fig:visual_2}.
Moreover, we observe that
the transformation function is capable of
either (1) gathering the positive and negative support nodes into two separate regions as shown in Fig.~\ref{fig:visual_1}
or (2) adjusting ``Query (+)'' and ``Query (-)'' to make the right prediction
when the positive and negative prototypes are close as shown in
Fig.~\ref{fig:visual_2}.
Another observation is that the transformation function has the tendency of enlarging the distances between node embeddings to facilitate classification.

\end{document}